\pdfoutput=1

\documentclass[11pt]{article}

\usepackage[final]{acl}

\usepackage{times}
\usepackage{pdfpages}
\usepackage{latexsym}
\usepackage{siunitx}
\usepackage{tabularx}
\usepackage[T1]{fontenc}
\usepackage[utf8]{inputenc}
\usepackage{CJKutf8}
\usepackage{csquotes}
\usepackage{microtype}
\usepackage{algorithm}
\usepackage{algpseudocode}
\usepackage{inconsolata}
\usepackage{adjustbox}
\usepackage{amsfonts} 
\usepackage{makecell}
\usepackage{graphicx} 
\usepackage{lscape}
\usepackage{multirow}
\usepackage{placeins}
\usepackage{listings}
\usepackage{tikz}
\usepackage{colortbl}
\usepackage{booktabs}
\usepackage{amsmath}
\usepackage{xspace}
\usepackage{float}
\usepackage[normalem]{ulem}
\newcommand{\stkout}[1]{\ifmmode\text{\sout{\ensuremath{#1}}}\else\sout{#1}\fi}

\title{CIG: Measuring Conversational Information Gain in Deliberative Dialogues with Semantic Memory Dynamics}  

\author{
Ming-Bin Chen \and Jey Han Lau \and Lea Frermann \\
School of Computing and Information Systems, The University of Melbourne \\
\texttt{\{mingbin,laujh,lfrermann\}@unimelb.edu.au}
}

\begin{document}
\maketitle

\begin{abstract}
Measuring the quality of public deliberation requires evaluating not only civility or argument structure, but also the informational progress of a conversation. We introduce a framework for \textbf{Conversational Information Gain (CIG)} that evaluates each utterance in terms of how it advances collective understanding of the target topic. To operationalize CIG, we model an evolving \textbf{semantic memory} of the discussion: the system extracts atomic claims from utterances and incrementally consolidates them into a structured memory state. Using this memory, we score each utterance along three interpretable dimensions: \textbf{Novelty}, \textbf{Relevance}, and \textbf{Implication Scope}. We annotate 80 segments from two moderated deliberative settings (TV debates and community discussions) with these dimensions and show that memory-derived dynamics (e.g., the number of claim updates) correlate more strongly with human-perceived CIG than traditional heuristics such as utterance length or TF--IDF. We develop effective LLM-based CIG predictors paving the way for information-focused conversation quality analysis in dialogues and deliberative success.\footnote{The code and annotation data are available at \url{https://github.com/mrknight21/memcig-analysis/tree/master/data}.}
\end{abstract}

\section{Introduction }\label{sec:intro}

\begin{figure*}[!t]
\centering
\includegraphics[width=0.95\textwidth]{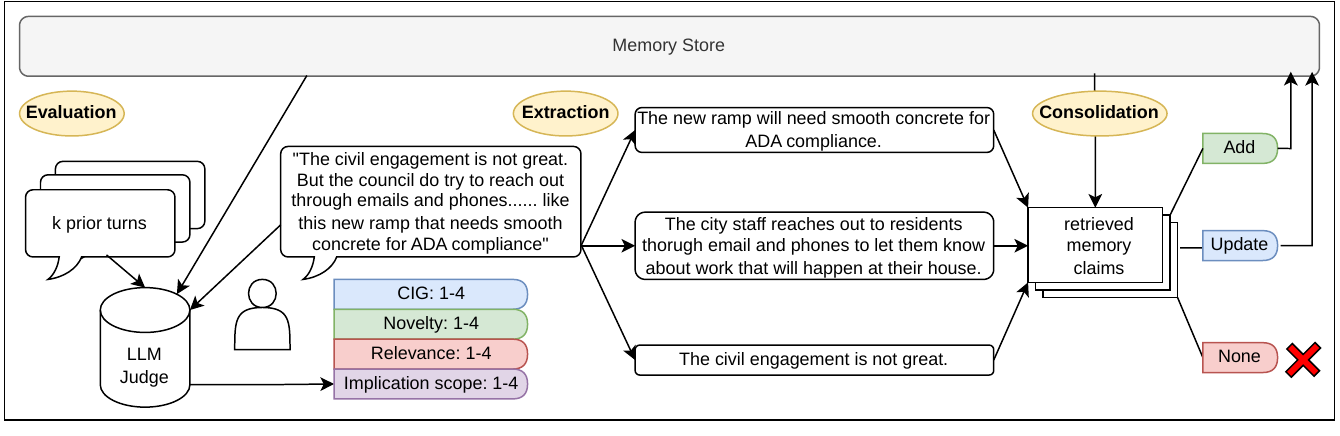}
\caption{
\textbf{Overview of the CIG pipeline.} 
Each utterance is \textbf{evaluated} with the Semantic Memory as knowledge context for Novelty, Relevance, Implication Scope, and overall CIG (1–4). 
The Semantic Memory is maintained through two modules: 
\textbf{Extraction}, which converts utterances into atomic claims; and 
\textbf{Consolidation}, which matches extracted claims against the retrieved memory, triggering \textsc{ADD}, \textsc{UPDATE}, or \textsc{NONE} operations.}
\label{fig:overall}
\end{figure*}

Public deliberation—reasoned dialogue aimed at collective understanding and decision-making~\citep{habermas1985theory} for public interest~\citep{dewey2012public}—is fundamental to democratic societies. Yet, the quality of these exchanges, from community forums \citep{schroeder2024fora} to public debates \citep{montez2019making}, can vary widely from stagnation to productive collaboration.

Metrics for evaluating dialogue quality \citep{goddard2023textual} have largely focused on structural formality over substantive content. Approaches based on schemes like the Deliberative Quality Index (DQI;  \citet{steenbergen2003measuring}) or computational proxies for civility \citep{anuchitanukul2022revisiting,price2020six} and argument structure \citep{wachsmuth2024argument} struggle to capture true informational progress. These surface-level signals can mislead: civil phrasing can conceal malicious intent \citep{kruk2024silent}, and valuable insights might emerge from informal exchanges \citep{walton2008informal, bohm2004dialogue}. Such methods fail to distinguish constructive progress from bureaucratic talk \citep{montez2019making, walton2003pragmatic}.

Grounded in Meadow and Yuan’s notion of information impact as \enquote{A change, or the nature or magnitude of change, in the knowledge base of a subject domain of the recipient} \citep[p.~710]{meadow1997measuring}, we define \textbf{Conversational Information Gain (CIG)} as the degree to which an utterance advances collective understanding toward the goal/topic. We decompose CIG into three interpretable aspects---\textbf{Novelty}, \textbf{Relevance}, and \textbf{Implication Scope}---capturing whether a contribution introduces new information, connects to the shared goal, and extends its implications to the public community beyond individual cases. To operationalize CIG, we require a representation of the evolving collective knowledge state against which each new utterance can be evaluated: both annotators and models must know what has already been said. As shown in Figure~\ref{fig:overall}, we implement this through a lightweight semantic memory that maintains a consolidated set of claims.

We first validate CIG through human annotation of 80 dialogue segments drawn from two moderated group-discussion settings—TV debates and community discussions—achieving moderate-to-high inter-annotator agreement for CIG and its aspects. We then validate automation by using an LLM with access to the same information as the human annotators (topic, short context, and a prior-memory summary), and show that LLM predictions closely track aggregated human judgments. Finally, by varying the model’s prior context, we find that predictions based on retrieved memory summaries are highly correlated with those based on the full preceding transcript, indicating that memory summaries provide a compact yet faithful substitute for full-history context in automated CIG assessment.

We then analyze how memory dynamics relate to perceived informational progress, finding that simple memory-state signals (e.g., claim update counts) track human CIG ratings more consistently than common heuristics such as sentence entropy or TF--IDF.
An unsupervised aggregation analysis reveals a ``conjunctive bottleneck'': an utterance’s CIG is effectively limited by its weakest aspect. Finally, we provide a case study illustrating how CIG can be used to analyze downstream interaction dynamics in moderated discussions.

\section{Related work}\label{sec:related-work}

\paragraph{Measuring Informativeness in Conversation}
While definitions of informativeness vary across disciplines, they share a common conceptual core: the change a message induces in the recipient—whether in certainty \citep{hunt2003concept}, utility \citep{glazer1993measuring}, or knowledge state \citep{meadow1997measuring}. Particularly, dialogue evaluation has largely converged on two necessary conditions required to trigger this change: novelty (the presence of new signal) and relevance (the alignment of that signal to the goal context)~\citep{ghosal2022novelty, maes2024did}. However, these two dimensions alone are insufficient to capture the \textit{magnitude} or \textit{worthiness} of a contribution~\citep{huang2020challenges}. While metrics like ``impact'' or ``usefulness'' attempt to proxy this third dimension, they are often vaguely defined or rely on subjective Likert scales that conflate personal preference with collective value~\citep{lee2022evaluating, qian2025bottom}. A complementary line of work extends classical information theory \citep{shannon1948mathematical}, leveraging surprisal or information-density to trace information flow \citep{maes2022shared, giulianelli2021information, tsipidi2024surprise}. However, such measures only partially align with human-perceived salience \citep{zarcone2016salience, maes2022shared}, limiting their precision in localizing information exchanges.

\paragraph{Agent Semantic Memory}
Agent memory modules provide a way to measure knowledge acquisition during conversation \citep{zhang2024survey}. Originally developed to sustain coherence and personalization in conversational agents over long conversation sessions, these modules function by maintaining a selective, persistent state. In frameworks such as Mem0 \citep{chhikara2025mem0}, salient claims are extracted from utterances and an LLM applies update operations to a knowledge store. This approach improves temporal and multi-hop reasoning on long-dialogue benchmarks \citep{maharana2024evaluating} and reduces latency and token cost relative to full-history baselines.

\begin{table*}[h]
\centering
\resizebox{\textwidth}{!}{
\begin{tabular}{p{0.2\linewidth} p{0.25\linewidth} p{0.9\linewidth}}
\toprule
\textbf{Aspect} & \textbf{Label} & \textbf{Definition / Anchor} \\ \midrule

\multirow{4}{1\linewidth}{Conversational Information Gain} 
& 1. No gain & Repeats or obstructs; no meaningful advance beyond the existing knowledge. \\
& 2. Minimal gain & Small clarification or slight nuance that is noticeable but limited. \\
& 3. Incremental & Adds new details/mechanisms within the same conceptual frame or ideas within the topic. \\
& 4. Insightful & Reframes or introduce new ideas under the topic; shifts the conversation in a new valuable way. \\ \midrule

\multirow{4}{1\linewidth}{Novelty} 
& 1. Not novel & Repetition/paraphrase of prior content or non/common-sense content. \\
& 2. Minimally novel & Minor or mostly predictable detail added to an existing idea. \\
& 3. Moderately novel & New evidence, concrete example, or supporting detail expanding an existing idea. \\
& 4. Highly novel & New framework, principle, idea, or line of reasoning that opens a new direction. \\ \midrule

\multirow{4}{1\linewidth}{Relevance} 
& 1. Not relevant & Off-topic; no connection to the conversation goal. \\
& 2. Minimally relevant & Loose or indirect link; requires inference to connect. \\
& 3. Moderately relevant & Substantially related but not central (e.g., side issue or counterpoint). \\
& 4. Highly relevant & Directly and explicitly addresses the core topic or goal. \\ \midrule

\multirow{4}{1\linewidth}{Implication Scope} 
& 1. Local & Manages the immediate moment; implication limited to participants/procedures. \\
& 2. Bounded & Self-contained fact, feeling, or stance; no generalization beyond the case. \\
& 3. Generalizing & Inductively generalizes a case or evidence to a broader audience. \\
& 4. Universal & States an abstract principle, value, or norm with wide or universal applicability. \\

\bottomrule
\end{tabular}
}
\caption{Conversational Information Gain (CIG) rubric. Each aspect is rated on a 1–4 scale using Prior Knowledge and the preceding dialogue as the collective knowledge context.}
\label{tab:cig_rubric_style}
\end{table*}

\section{Conversational Information Gain (CIG)}

Following previous studies in group discussion analysis \cite{dowell2019group}, we define \textbf{Conversational Information Gain (CIG)} as \textit{how much a response advances the group’s shared understanding of the topic or progress toward the goal, given both prior knowledge and the preceding dialogue}. We decompose CIG into three aspects—\textbf{Novelty}, \textbf{Relevance}, and \textbf{Implication Scope}—to connect general criteria for informativeness to the demands of public deliberation. Novelty and Relevance capture whether a contribution introduces new information and is tied to the discussion goal, while Implication Scope captures how broadly the contribution’s implications extend beyond the immediate case and people, reflecting the public orientation of deliberation. All constructs are rated on a four-level scale (Table~\ref{tab:cig_rubric_style}). Importantly, CIG reflects informational advancement rather than conversational quality per se—reiterations or coordination moves typically score low on CIG, even though they may still support the discussion indirectly.

To anchor high CIG levels, we draw on \citet{chi2009three}'s typology of conceptual change which distinguishes \textit{assimilation}—adding information within an existing mental model—from \textit{accommodation}—restructuring the model itself. Accordingly, we separate our top two CIG levels: Level~3 (\textit{Incremental}) reflects assimilation, where an utterance adds evidence, details, or mechanisms within the current framing; Level~4 (\textit{Insightful}) reflects accommodation, where an utterance introduces a new framing or principle that qualitatively redirects the discussion. Detailed definitions for each aspect are in Table~\ref{tab:cig_rubric_style}, and examples in Appendix Table~\ref{tab:cig_rubric_example}.

\paragraph{Novelty} \textit{Assesses whether the information is new compared to the prior knowledge and preceding dialogue}. Traditional approaches often proxy novelty with n-gram overlap against prior text \citep{soboroff2005novelty}, but recent work shows that lexical novelty correlates poorly with human perception \citep{saakyan2025death}. Consequently, we rate novelty on a similar four-level scale as CIG, with levels reflecting the message’s effect on the existing conceptual framing. The novelty score is intentionally independent of topical alignment and magnitude of impact.

\paragraph{Relevance} \textit{Measures how substantively a message relates to the main conversation topic or goal.} Since lexical-overlap proxies correlate weakly with human judgments in context-sensitive and implicit dialogue \citep{yeh2021comprehensive,dascal1977conversational}, we adopt the four-level topic--conversation relevance scale \citep{fan2024topic}. Levels~1 (off-topic) and~4 (directly on-topic) form the endpoints; the key distinction is between the middle tiers. Level~2 (\textit{minimally relevant}) covers content whose connection to the goal is indirect and requires a bridging inference (e.g., “school choice” for topic “housing affordability”), and Level 3 covers content that is clearly connected but not central, typically a recognized subtopic (e.g., “zoning restrictions”).

\paragraph{Implication Scope}
\textit{Measures the intended reach of a statement—who it is meant to matter to and how far its implications generalize.} Many dialogue-evaluation schemes include an ``impact'' dimension \citep{lee2022evaluating}. In public deliberation, however, perceived impact is subjective and highly context-dependent. We therefore operationalize this idea—drawing on the notion of ``generality'' in \citet{meadow1997measuring}—as the \emph{scope of implication} a contribution projects beyond the immediate speaker and case. This is deliberation-motivated: participants often move between \emph{situated testimony} (what happened to me/us here) and \emph{public reasons} (what should matter to the community), and this ``lifting'' from particulars to publicly addressable concerns is commonly treated as central to deliberative legitimacy \citep{habermas1985theory}.

Our rubric has four levels. Level~1 covers local or procedural moves whose significance is confined to the immediate interaction (e.g., turn management). Level~2 captures bounded, case-specific content (e.g., a personal fact or stance) that does not generalize beyond the case. Level~3 reflects inductive generalization, where a specific experience or piece of evidence is used to motivate a broader pattern—such as testimony offered to evidence social harm \citep{kessler2023hearing}. Level~4 denotes universal, principle-level claims framed for wide public applicability (e.g., rights, fairness, norms). We emphasize that this hierarchy is \emph{descriptive rather than normative}: effective deliberation often requires a rhythm between grounded particulars (Levels~2--3) and abstract principles (Level~4). Accordingly, higher scope does not necessarily imply greater \emph{informativeness}; rather, Scope characterizes \emph{how} a contribution frames its relevance to the public, which we later test as a potential (but not guaranteed) driver of perceived CIG.

\section{Entailment-based Semantic Memory}

To automatically and incrementally score the CIG of conversational utterances, we require a dynamic representation of what the conversation has already established. To this end, we extend the Mem0 framework \citep{chhikara2025mem0} to a multi-party setting, maintaining an evolving semantic memory that tracks how each speaker introduces and revises claims over time. As shown in Figure \ref{fig:overall}, the memory is managed through a two-stage process of claim extraction and consolidation to provide the knowledge context for evaluating Novelty, Relevance, Implication Scope, and overall CIG.

\paragraph{Claim Extraction}
The first stage uses an LLM to decompose each utterance into a set of discrete, self-contained claims, converting context-dependent dialogue into a structured representation for semantic memory consolidation and storage (Appendix Table~\ref{tab:claim_extraction_prompt} details the prompt). To make each claim interpretable in isolation, the extractor uses local context to resolve ambiguities (e.g., coreference) and splits compound statements into atomic propositions, while normalizing away conversational fillers and hedging to recover the core propositional content (e.g., ``I feel that maybe the transport cost is too high'' $\rightarrow$ ``The transport cost is too high''). Although this normalization removes epistemic markers, it enables more reliable NLI-based entailment checks; we preserve speaker-specific beliefs by storing each extracted claim together with its source speaker and inferred addressee in the memory state.

\paragraph{Multi-Party Memory Consolidation}

To manage redundancy and track knowledge evolution, each newly extracted claim $A$ is compared against the semantic memory by first retrieving the \mbox{top-$k$} most semantically similar candidates $\{B_i\}_{i=1}^{k}$ via vector search. An LLM-based NLI judge bi-directionally evaluates the claim and each candidate, mapping each pair to one of five relations: \textit{equivalent}, \textit{forward\_entail}, \textit{backward\_entail}, \textit{contradiction}, or \textit{neutral}.

We then apply a deterministic policy to select an action $\alpha(A)\in\{\textsc{ADD}, \textsc{UPDATE}, \textsc{NONE}\}$. To strictly track speaker-specific belief trajectories without collapsing inter-speaker disagreements, we restrict operations to the same-speaker subset $\mathcal{S}(A) = \{B_i \mid \text{speaker}(B_i)=\text{speaker}(A)\}$. If $\mathcal{S}(A)=\emptyset$ or contains only \textit{neutral} relations, we trigger \textsc{ADD} and insert $A$ into the semantic memory; this ensures that parallel framings or contested claims from other participants are preserved as distinct entries. For $\mathcal{S}(A)\neq\emptyset$, \textit{equivalence} or \textit{backward\_entailment} triggers \textsc{NONE} (taking priority over updates), while \textit{contradiction} or \textit{forward\_entailment} triggers \mbox{\textsc{UPDATE}} on the single most similar item $B^\star=\arg\max_{B\in\mathcal{S}(A)}\text{sim}(A,B)$. Here, \textsc{UPDATE} means revising the existing memory $B^\star$ using the new claim $A$: contradictions replace $B^\star$ with $A$, whereas forward entailments merge $A$ and $B^\star$ into a more specific consolidated claim. Appendix Table~\ref{tab:nli-to-action-same} summarizes this mapping with examples, and Appendix Table~\ref{tab:memory_consolidation_prompt} details the prompt.

\section{Data and Annotation}

\begin{table}[t]
\centering
\small
\setlength{\tabcolsep}{3pt}
\begin{tabular}{lcc}
\toprule
& \textbf{FORA} & \textbf{INSQ} \\
\midrule
\multicolumn{3}{l}{\textit{Session-level (10 per setting, avg $\pm$ std)}} \\
Utterances / Session & 210.1 $\pm$ 56.5 & 205.8 $\pm$ 73.8 \\
Words / Utterance & 38.8 $\pm$ 11.9 & 36.4 $\pm$ 17.9 \\
Speakers / Session & 6.8 $\pm$ 1.0 & 8.8 $\pm$ 1.8 \\
\midrule
\multicolumn{3}{l}{\textit{Annotated segments (4 per setting, avg $\pm$ std)}} \\
Utterances / Segment & 11.6 $\pm$ 4.4 & 11.3 $\pm$ 4.7 \\
Speakers / Segment & 4.5 $\pm$ 1.3 & 4.4 $\pm$ 1.0 \\
Words / Non-skipped Utterance & 86.3 $\pm$ 52.5 & 88.6 $\pm$ 56.0 \\
Speaker Gini & 0.42 $\pm$ 0.16 & 0.21 $\pm$ 0.15 \\
Skipped Tokens (\%) & 17.8 $\pm$ 21.0 & 9.4 $\pm$ 9.0 \\
\bottomrule
\end{tabular}
\caption{
Session- and segment-level descriptive statistics for the annotated \textit{FORA} and \textit{INSQ} dialogues. For segment-level statistics, moderator turns and short fragmentary utterances marked as \textit{skipped} are excluded when computing \textit{Words / Non-skipped Utterance} and \textit{Speaker Gini}. The speaker-participation Gini coefficient measures inequality in contribution within a segment, with higher values indicating more uneven participation.
}
\label{tab:segstats_revised}
\end{table}

We annotated transcripts from two deliberative discussion settings: ten TV debates from \textbf{\textsc{Intelligence Squared}} ({\bf INSQ};~\citet{zhang2016conversational}) and ten community discussions from the \textbf{FORA corpus} ({\bf FORA};~\citet{schroeder2024fora}) about Durham's community future vision. To mitigate annotator bias and potential harms, we manually curated this set by excluding topics of a highly sensitive or recently polarizing nature. While both settings feature a moderator and multi-party discussion (see Table~\ref{tab:segstats_revised} for descriptive statistics), they differ in atmosphere and intent. \textsc{INSQ} debates are competitive events organized around a motion and two teams, featuring domain experts in which the moderator primarily coordinates turn-taking. \textsc{FORA} assemblies are collaborative, brainstorming meetings where facilitators often participate as peers alongside general community members. This contrast is also visible in the speaker Gini values in Table~\ref{tab:segstats_revised}: \textsc{INSQ} shows more even participation, consistent with tighter moderation, whereas \textsc{FORA} is more unequal, reflecting a freer and more community-driven discussion style.

\subsection{Preprocessing}
Both FORA and INSQ exhibit a comparable three-part format—introduction, discussion, and conclusion. 
For \textsc{INSQ}, we use the dataset’s corresponding predefined structure and analyze only the discussion phase. In \textsc{FORA} the boundaries are not annotated, so we use GPT-5~\citep{gpt5_api} to detect them and manually verify. In both settings, \textit{conclusions} are discarded, and \textit{introductions} are used to initialize the initial knowledge in the memory modules. We then segment the \textit{discussion} phase of each episode into sub-topical units using \mbox{GPT-5} with a shifting window and majority vote (see Appendix~\ref{appendix:segment} for segmentation details).

For each segment, we generate a prior memory summary by retrieving relevant existing memories via semantic similarity and summarizing them using GPT-5. We developed the summarization prompt using pilot data annotated by one paper author (see Appendix~\ref{sec:memory_summarisation} for development details). To focus the annotation on participants' contributions, moderator turns and short utterances (\(<5\) words) (e.g. ``I want to-.'') are marked as \textit{skipped}, accounting for 9--18\% of tokens (see Table~\ref{tab:segstats_revised}). From each episode, we select four segments for annotation, balancing factors like estimated reading time, skip ratio, and segmentation confidence (see Appendix~\ref{sec:segment_selection} for the selection criteria).

\begin{table}[t]
\centering
\footnotesize
\setlength{\tabcolsep}{0pt}
\begin{tabular*}{\columnwidth}{@{\extracolsep{\fill}}l c c c c}
\toprule
     & \textbf{CIG} & \textbf{Novelty} & \textbf{Relevance} & \textbf{Scope} \\ 
\midrule
INSQ & 0.589        & 0.506            & 0.669              & 0.597          \\
FORA & 0.567        & 0.583            & 0.566              & 0.510          \\ 
\bottomrule
\end{tabular*}
\caption{Krippendorff’s $\alpha$ by corpus for CIG (Informativeness) and its three aspects.}
\label{tab:kpf}
\end{table}

\subsection{Annotation Protocol}
We recruited 88 annotators on Prolific (\pounds{}10.60/hour). All participants were required to be native English speakers with at least a bachelor's degree. We ultimately dropped 10 participants due to consistently low agreement with peers.\footnote{Our quality control process involved recruiting two annotators per session first. If their mean quadratic weighted kappa (QWK) score was ${<}0.25$, we recruited a third annotator. We then identified and removed the outlier annotator, ensuring at least two reliable annotations per segment.} Each assignment (approximately 40 minutes) included a tutorial and a prescreen test, followed by four sampled segments from the same session presented in chronological order. 

We deployed two annotation variants over the same material: one collected a single score for overall CIG per utterance (on a scale of 1--4); the other collected three separate ratings for \textit{Novelty}, \textit{Relevance}, and \textit{Implication Scope}, each on a scale of 1--4. The annotation interface showed the topic at the top, with long-term memory (the prior memories summary) and short-term context (three preceding utterances) in a left panel, and the current target utterance for rating in a right panel (see Appendix~\ref{sec:annotation_interface} for the interface snippet and design). Inter-annotator agreement, reported in Table~\ref{tab:kpf}, is moderate to high across all aspects and settings. Appendix Figure~\ref{fig:label-distributions} shows the normalized label distributions.

\section{Validating the CIG Aspects}

To test whether overall CIG is explained by the three proposed aspects, we fit ordinal regression models on the 80 annotated segments and compare different predictor sets against a word-count baseline (Table~\ref{tab:ord_models_compact}). AIC-based model selection suggests that Novelty and Relevance account for most of the predictive signal. In contrast, Implication Scope---which we included to capture a deliberation-specific intuition that contributions with broader public reach might be perceived as more informative---provides little reliable benefit. In \textsc{INSQ}, the best fit is achieved by removing Scope, indicating that including it slightly degrades predictive power; in \textsc{FORA}, the full model improves only marginally over the Scope ablation.

Novelty and Relevance behave as expected and explain most of the variance in perceived CIG. In contrast, Implication Scope contributes little. This challenges our deliberation-motivated hypothesis that statements framed with broader public reach would systematically be judged as more informative than local or case-specific remarks. One plausible explanation is that informativeness is sensitive to \emph{grounding} and \emph{evidential value}, not only generality: low-scope utterances can be highly informative when they introduce concrete facts or lived experiences that audiences treat as diagnostic, whereas high-scope utterances can be uninformative when they restate abstract values without new content. This interpretation aligns with findings that hearing personal experiences can improve social evaluations compared to hearing personal opinions, particularly under polarization, suggesting that situated testimony can carry distinctive communicative value even without being framed as universal public principle~\citep{kessler2023hearing}. Accordingly, Scope may be better interpreted as a descriptive axis of deliberative style rather than a monotonic driver of perceived CIG.

\begin{table}[t]
\centering
\footnotesize
\setlength{\tabcolsep}{0pt} 
\begin{tabular*}{\columnwidth}{@{\extracolsep{\fill}}l l r@{}}
\toprule
\textbf{Corpus} & \textbf{Model} & \textbf{AIC} ($\downarrow$) \\
\midrule
\multirow{7}{*}{INSQ}
  & Base (word count)       & 472.58 \\
  & Base+novelty               & 424.39 \\
  & Base+relevance               & 444.29 \\
  & Base+implication scope               & 445.92 \\
  & Base+3 (all aspects)    & 416.60 \\
  & \textbf{Ablation(-implication scope)} & \textbf{416.01} \\
  & Ablation(-relevance)         & 420.21 \\
  & Ablation(-novelty)         & 438.26 \\
\midrule
\multirow{7}{*}{FORA}
  & Base (word count)       & 643.25 \\
  & Base+novelty               & 599.28 \\
  & Base+relevance               & 580.12 \\
  & Base+implication scope               & 637.83 \\
  & \textbf{Base+3 (all aspects)} & \textbf{564.06} \\
  & Ablation(-implication scope)         & 564.53 \\
  & Ablation(-relevance)         & 601.09 \\
  & Ablation(-novelty)         & 581.83 \\
\bottomrule
\end{tabular*}
\caption{
Ordinal regression model comparison using AIC (lower is better). 
Base uses only utterance length. Base+3 includes all three aspects.
Ablation models remove one aspect from the full model.
}
\label{tab:ord_models_compact}
\end{table}

\section{Prediction of CIG and Heuristic Aggregation}

\begin{table}[t]
\centering
\footnotesize
\setlength{\tabcolsep}{5pt}
\begin{tabular}{l l cc}
\toprule
\textbf{Corpus} & \textbf{Aspect} 
& \textbf{GPT-5} 
& \textbf{Human LOO} \\
\midrule
\multirow{4}{*}{INSQ}
  & CIG         & 0.457$\pm$0.020 & 0.656$\pm$0.131 \\
  & Novelty     & 0.587$\pm$0.024 & 0.637$\pm$0.245 \\
  & Relevance   & 0.452$\pm$0.018 & 0.431$\pm$0.202 \\
  & Imp.~Scope  & 0.529$\pm$0.021 & 0.562$\pm$0.162 \\
\midrule
\multirow{4}{*}{FORA}
  & CIG         & 0.520$\pm$0.027 & 0.631$\pm$0.141 \\
  & Novelty     & 0.556$\pm$0.017 & 0.599$\pm$0.200 \\
  & Relevance   & 0.414$\pm$0.015 & 0.446$\pm$0.163 \\
  & Imp.~Scope  & 0.479$\pm$0.008 & 0.559$\pm$0.226 \\
\bottomrule
\end{tabular}
\caption{
Mean absolute error for \textbf{GPT-5} when given only the memory summary—the same information provided to human annotators—compared to the \textbf{Human LOO} baseline.
}
\label{tab:gpt-summary-vs-human}
\end{table}

\begin{table*}[t]
\centering
\footnotesize
\setlength{\tabcolsep}{6pt}
\resizebox{\linewidth}{!}{
\begin{tabular}{l l cccc}
\toprule
\textbf{Corpus} & \textbf{Aspect} 
& \textbf{GPT-5\_full} 
& \textbf{GPT-5\_memory} 
& \textbf{GPT-5\_short\_prior} 
& \textbf{GPT-5\_no\_knowledge} \\
\midrule
\multirow{5}{*}{INSQ}
  & CIG        & 0.265$\pm$0.021 & \textbf{0.265}$\pm$0.007 & 0.350$\pm$0.013 & 0.394$\pm$0.024 \\
  & Novelty    & 0.353$\pm$0.009 & \textbf{0.321}$\pm$0.028 & 0.649$\pm$0.038 & 0.764$\pm$0.044 \\
  & Relevance  & 0.159$\pm$0.026 & 0.131$\pm$0.013 & \textbf{0.125}$\pm$0.007 & 0.151$\pm$0.025 \\
  & Imp.~Scope & 0.165$\pm$0.011 & 0.156$\pm$0.014 & \textbf{0.149}$\pm$0.011 & 0.155$\pm$0.016 \\
  & \textbf{Mean} & 0.236$\pm$0.017 & \textbf{0.218}$\pm$0.016 & 0.318$\pm$0.017 & 0.366$\pm$0.027 \\
\midrule
\multirow{5}{*}{FORA}
  & CIG        & \textbf{0.207}$\pm$0.008 & 0.220$\pm$0.010 & 0.293$\pm$0.020 & 0.336$\pm$0.008 \\
  & Novelty    & 0.279$\pm$0.006 & \textbf{0.259}$\pm$0.019 & 0.383$\pm$0.031 & 0.492$\pm$0.025 \\
  & Relevance  & 0.151$\pm$0.015 & \textbf{0.146}$\pm$0.012 & 0.148$\pm$0.006 & 0.167$\pm$0.005 \\
  & Imp.~Scope & 0.211$\pm$0.027 & 0.203$\pm$0.007 & \textbf{0.195}$\pm$0.010 & 0.206$\pm$0.010 \\
  & \textbf{Mean} & 0.212$\pm$0.014 & \textbf{0.207}$\pm$0.012 & 0.255$\pm$0.017 & 0.300$\pm$0.012 \\
\midrule
\multicolumn{2}{l}{\textbf{Overall mean (both corpora)}} 
& 0.224$\pm$0.015 & \textbf{0.213}$\pm$0.014 & 0.286$\pm$0.017 & 0.333$\pm$0.020 \\
\bottomrule
\end{tabular}
}
\caption{
MAE of GPT-5 variants evaluated against \textbf{GPT-5\_summary} reference, across both corpora and all aspects. Bold indicates the best performer across all variants.}
\label{tab:gpt-mae-summary-ref}
\end{table*}

\paragraph{Predicting CIG with GPT-5}
We next test whether segment-level CIG annotation can be automated with GPT-5 under the same information conditions as our human annotators. For each target segment, GPT-5 receives (i) the memory-based prior summary, (ii) the three preceding utterances, and (iii) the  tutorial exemplars presented to annotators as few-shot demonstrations, and outputs ratings (Table~\ref{tab:cig_rubric_style}) for overall CIG and its aspects (Appendix Tables~\ref{tab:rating_prompt_info} and~\ref{tab:rating_prompt_mix} detail the prompts). To contextualize performance, we report a human leave-one-out (LOO) baseline, computed as each annotator’s MAE against the mean of the remaining annotators and averaged across annotators, which estimates the expected deviation of an individual annotator from group consensus. As shown in Table~\ref{tab:gpt-summary-vs-human}, GPT-5’s MAE under this matched-context setup is comparable to—and in several cases lower than—human LOO error, suggesting that GPT-5 can reproduce aggregate human judgments at least as well as a typical annotator when constrained to the same prior context (additional models are reported in Appendix Table~\ref{tab:mae_results}).

We next examine how alternative prior-knowledge inputs reproduce the \textsc{GPT-5\_summary} predictions, to check whether the memory-derived summary retains the information that most influences GPT-5’s ratings. Specifically, we compare four conditions: the full preceding transcript (\texttt{GPT-5\_full}), the retrieved memory items shown verbatim without summarisation (\texttt{GPT-5\_memory}), only the three most recent utterances (\texttt{GPT-5\_short\_prior}), and no prior context beyond the topic (\texttt{GPT-5\_no\_knowledge}). As shown in Table~\ref{tab:gpt-mae-summary-ref}, \texttt{GPT-5\_full} and \texttt{GPT-5\_memory} yield the smallest mean deviations from \textsc{GPT-5\_summary}, while restricting or removing prior context produces substantially larger errors. Overall, this suggests that the memory-based summaries provide a compact yet faithful approximation of the information in the full history that drives GPT-5’s CIG estimates.

\paragraph{Comparing memory-dynamics signals to heuristic proxies}
We compare several informativeness-related heuristic proxies used in prior work—word-based, surprisal-based, entity-based, and TF--IDF features—against our memory-dynamics signals (per-utterance counts of extracted claims and claim updates) by computing Pearson correlations with the human-annotated CIG soft label (mean rating across annotators). As shown in Table~\ref{tab:feat_corr_human}, memory dynamics are the strongest correlates: relevance-gated updates (\textit{Memory changes (Relv\textsuperscript{+})}) achieve the highest correlation ($|r|=0.726$), followed by \textit{Memory changes (Any)} ($|r|=0.720$) and \textit{Extracted claim count} ($|r|=0.713$), outperforming all other proxies. \textit{Token surprisal sum} and utterance \textit{Length (tokens)} form a mid-tier baseline ($|r|=0.678$ and $|r|=0.697$, respectively), while entity-based signals are substantially weaker (e.g., \textit{Entity count} $|r|=0.466$). Because annotators were provided memory-derived prior context, memory dynamics may be advantaged; we therefore replicate the same correlation analysis using GPT-5 predictions produced under full-transcript context (\textsc{GPT-5\_full}) and observe a similar overall pattern (Appendix Table~\ref{tab:feat_corr_gpt5}). Appendix~\ref{appendix:proxy_equations} details the proxy definitions.

\begin{table}[t]
\centering
\small
\setlength{\tabcolsep}{6pt}
\renewcommand{\arraystretch}{1.15}
\begin{tabular}{l c}
\toprule
\textbf{Feature} & \textbf{$|r|$ w/ \texttt{CIG}} \\
\midrule
Memory changes (Relv\textsuperscript{+}) & 0.726 \\
Memory changes (Any) & 0.720 \\
Extracted claim count & 0.713 \\
Memory changes (Info\textsuperscript{+}) & 0.712 \\
Memory changes (Novo\textsuperscript{+}) & 0.703 \\
TF--IDF sum & 0.701 \\
Length (tokens) & 0.697 \\
Token surprisal sum & 0.678 \\
TF--IDF max & 0.638 \\
Novel word count & 0.601 \\
Memory changes (Scope\textsuperscript{+}) & 0.592 \\
Entity count & 0.466 \\
Novel entity count & 0.419 \\
Novel entity ratio & 0.298 \\
Specificity (mean IDF) & 0.223 \\
\bottomrule
\end{tabular}
\caption{Pearson correlation between each proxy feature and the \texttt{CIG} average from \textbf{human} annotations. Aspect-gated memory-change features (marked with \textsuperscript{+}) count updates only for utterances whose corresponding aspect score is greater than 2.}
\label{tab:feat_corr_human}
\end{table}

\paragraph{Claim-Level Predictions to Utterance Impression}
Motivated by broader interest in semantic salience \citep[e.g.,][]{biggs2012semantic}, we conduct an exploratory analysis to examine how claim-level qualities roll up to utterance-level impressions of CIG \citep{wei2022investigating}. First, we use GPT-5 to predict scores (1--4) for every extracted claim on all three aspects (Novelty, Relevance, and Scope; see Appendix Table~\ref{tab:claim_rating_prompt_full} for prompt details). We then apply a two-step, unsupervised procedure to test how well these claim-level predictions can recover human-annotated utterance-level CIG scores. This involves: (i) aggregating aspect-specific scores across multiple claims within an utterance (claim-aggregation on the x-axis), and (ii) combining the resulting aspect scores into a single CIG estimate (aspect-combination on the y-axis). We test a range of operators for both steps.

\begin{figure}[t]
  \centering
  \includegraphics[width=\columnwidth]{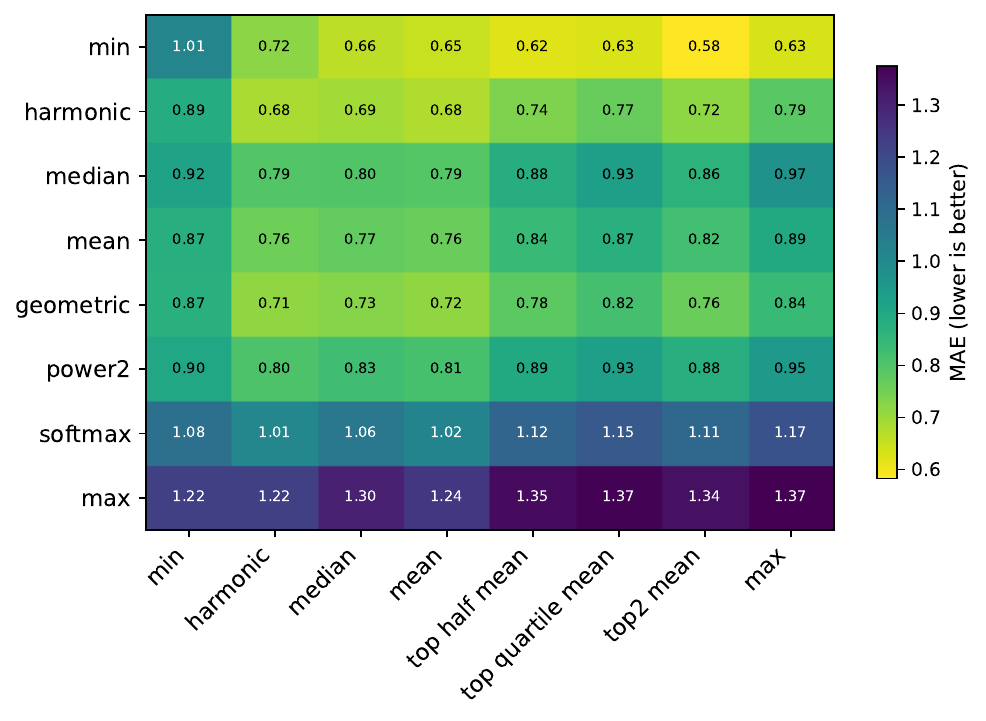}
  \caption{Heatmap of MAE for predicting human utterance-level CIG. The \textbf{y-axis} represents \textbf{Aspect Combination methods}. The \textbf{x-axis} represents \textbf{Claim Aggregation methods}.}
\label{fig:agg_mae_heatmap}
\end{figure}

The results, shown in Figure~\ref{fig:agg_mae_heatmap}, reveal two consistent patterns. First, for aspect combination (y-axis), the ``min'' operator performs best by a large margin across all claim-level aggregators. This suggests a \textbf{conjunctive bottleneck}: utterance-level informativeness is effectively constrained by the weakest aspect, rather than increasing smoothly with the average strength of all aspects. Second, for claim aggregation (x-axis), the best-performing operators are selective pooling methods such as ``top-2 mean'' (which achieves the lowest MAE of 0.583) and ``top-quartile mean'' (MAE = 0.627), especially when paired with ``min''. By contrast, combination operators that are more sensitive to high values, such as ``softmax'' and ``max'', perform worst. Taken together, these findings suggest that listeners do not judge an utterance by uniformly averaging all of its claims. Instead, impressions of CIG appear to be driven by a small number of salient claims, while still requiring that no core aspect be critically weak. This supports our broader view that perceived informativeness in dialogue is multi-dimensional and bottlenecked, rather than a simple additive function of semantic content.

\section{Case Study: Moderator Dynamics}\label{sec:mod}

\begin{figure}[h]
\centering
\includegraphics[width=\columnwidth]{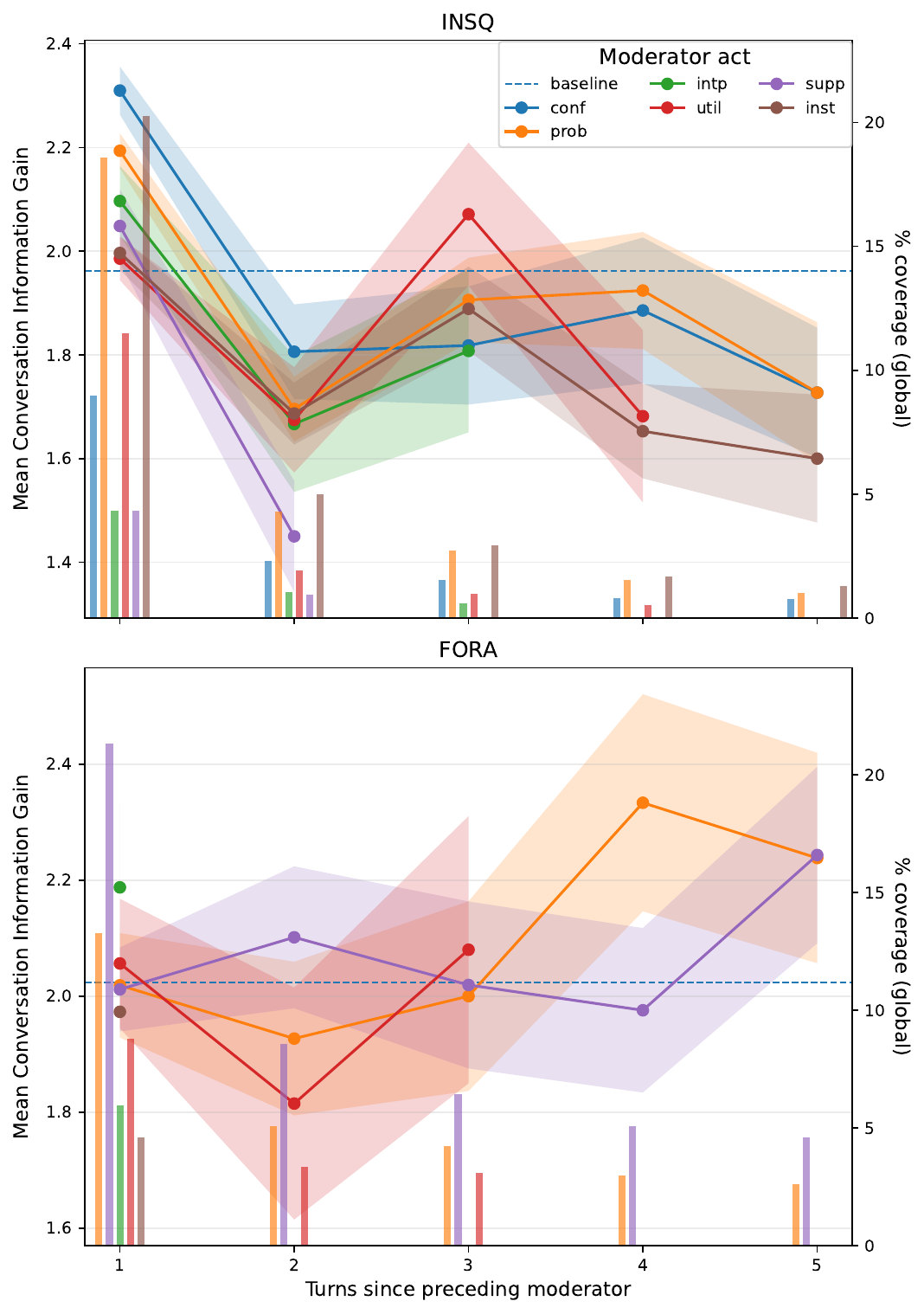}
\caption{Mean participant CIG (left y-axis) vs. turns since the last moderator act (x-axis), for both settings. Lines show the mean CIG for each moderator act. Bars (right y-axis) show the global \% coverage for each act. The dashed line is the corpus mean CIG.}
\label{fig:mod_lag}
\end{figure}

To demonstrate CIG's utility beyond static evaluation, we examine whether it can diagnose interaction dynamics—specifically, how moderator interventions shape subsequent informational progress. Facilitators play a central role in deliberation, yet their effects are usually described qualitatively or through broad outcome measures rather than through the local informational gain they elicit. To categorize these interventions, we employ the \textbf{WHoW taxonomy} \citep{chen2024whow}, a domain-agnostic framework designed to analyze moderator behavior across varying contexts. WHoW classifies moderator turns into six broad communicative acts: \textit{Probing} (eliciting responses), \textit{Confronting} (direct engagement between participants), \textit{Interpretation} (interpreting previous content), \textit{Supplementing} (contributing information), and \textit{Utilities} (other functional acts). By mapping downstream CIG trajectories to these specific acts, we can isolate which strategies trigger informational gain and how these effects vary by context.

To quantify these dynamics, we align each participant utterance to the most recent moderator turn. These pairs are grouped by moderator act type and indexed by their temporal lag—specifically, the number of turns (1–5) following the intervention. We then compute the mean participant CIG at each lag for every act type while accounting for the overall prevalence of each act within the corpus.

As shown in Figure~\ref{fig:mod_lag}, a clear divergence emerges between the two settings. In \textsc{INSQ}, moderator interventions are followed by sharp, immediate CIG peaks, with the strongest gains typically occurring at lag 1. This is particularly evident in \textit{Confronting} acts, which align with the adversarial debate format where moderators actively sharpen disagreement to prompt direct rebuttals. In \textsc{FORA}, by contrast, peaks are more diffused and often emerge several turns later. This suggests that in community discussions, informational progress is less tethered to a single facilitator move, instead accumulating gradually through a sequence of follow-on contributions. Notably, \textit{Confronting} acts are absent in \textsc{FORA}, consistent with its less adversarial and more facilitative interaction style.

This contrast shows that CIG is sensitive not only to the content of individual utterances but also to the interactional regime in which they are produced. In \textsc{INSQ}, moderators manage turn-taking and actively challenge participants within a competitive structure, so interventions often elicit immediate high-information responses. In \textsc{FORA}, facilitation is looser and more participatory, with more \textit{Supplementing} moves from facilitators themselves, so informational progress appears more distributed across subsequent turns. More broadly, this case study illustrates how CIG can serve as a diagnostic lens for comparing moderation styles, revealing not only whether a discussion is informative overall, but also how information gain is locally elicited and sustained over time.

\section{Conclusions}\label{sec:conclusions}

We introduced CIG, a framework for measuring \textit{Conversational Information Gain} in deliberative dialogue by focusing on informational progress rather than surface form. Grounded in the notion of information gain as a change in a shared knowledge state~\citep{meadow1997measuring}, our approach operationalizes CIG through a lightweight semantic memory maintained via LLM-based claim extraction and NLI-guided consolidation. This evolving record of claims provides the necessary context to evaluate utterances along four interpretable dimensions: overall CIG, Novelty, Relevance, and Implication Scope.

Across two contrasting deliberative settings—\textsc{INSQ} debates and \textsc{FORA} community discussions—we found that CIG can be annotated with moderate-to-high reliability and meaningfully decomposed into interpretable aspects. Novelty and Relevance emerged as the primary drivers of perceived CIG, while Implication Scope was more informative as a descriptive dimension of deliberative style than as a consistent predictor of informativeness. GPT-5 was able to reproduce aggregate human judgments under matched information conditions, suggesting that CIG can be assessed automatically with reasonable fidelity.

From a broader perspective, these results highlight the value of tracking semantic memory dynamics directly. Memory-based signals—such as counts of extracted claims and claim updates—aligned more strongly with human-perceived CIG than common heuristic proxies. Our analysis further suggests that informativeness perception is governed by a conjunctive bottleneck: an utterance is only perceived as informative as its weakest dimension allows. Finally, the moderation case study demonstrates CIG’s utility as a diagnostic lens. In \textsc{INSQ}, moderator interventions catalyzed immediate informational surges, whereas in \textsc{FORA}, gains accumulated more gradually across subsequent turns. Together, these findings position CIG as a framework for studying how information is introduced, elicited, and sustained in deliberative interaction.

Future work could integrate newer memory architectures~\citep{xu2025mem, li2025memos} to improve consolidation robustness and long-context tracking. The NLI-based update mechanism may also support analyses of finer-grained meaning construction in multi-party dialogue~\citep{poole2025ai}. Furthermore, memory-based information gain signals may provide interpretable supervision or even reward signals for training and evaluating information-seeking, deliberation-support, and facilitation-oriented conversational systems.

\section*{Limitations}\label{sec:limitations}
Our study has several limitations. First, the CIG labels are collected under a constrained information condition (topic, a prior-knowledge summary, and a short preceding window), which may bias judgments toward what is captured by the summary rather than the full discourse; while we partially probe this with context ablations, the results may not generalize to settings where annotators have full access to long histories. Second, the semantic-memory pipeline relies on LLM extraction and NLI-style consolidation, so errors in claim parsing, retrieval, or entailment can propagate into memory dynamics and any downstream analyses. Third, due to efficiency constraints, each extracted claim is assigned only a single memory action, and at most one relation to one existing memory item is recorded, which can under-represent cases where a claim simultaneously relates to multiple prior claims. Finally, we study two moderated English-language corpora and exclude highly sensitive topics; broader coverage (languages, cultures, and adversarial/polarized domains) is needed to assess robustness and external validity.

\section*{Ethical considerations}\label{sec:ethics}
We recruited annotators via an online platform to rate the informativeness of dialogue segments drawn from public deliberative settings. Participation was voluntary and based on informed consent; annotators were informed of their right to withdraw at any time without penalty and could opt out of rating specific segments after viewing the topic if they felt uncomfortable(Appendix~\ref{appendix:pls} show anonymised version of the plain language statement for participation). Because public discussions can include sensitive or emotional content, we limited topic selection and provided sufficient context to support informed skipping. We collected no direct personal identifiers beyond platform-managed participant IDs, and we release only de-identified annotations. Compensation followed platform guidelines and was set to meet or exceed applicable minimum-wage expectations for estimated task duration. The annotation protocol and material were approved by the University of Melbourne research ethics committee with the reference code- 2023-28400-47354-1. We also acknowledged that the studies have used large language model, including Gemini-2.5-pro from Google and GPt-5 from OpenAI, for code refactoring and data cleaning.

\section*{Acknowledgment}

 We thank Hope Schroeder for providing access to the FORA corpus, and Yin Ma for guidance on using Prolific and advice on annotation interface design. We also thank Rena Gao, Zheng Wei Lim, and Aso Mahmudi for participating in pilot studies. We are grateful to members of the Center for Constructive Communication at MIT Media Lab for their inspiration during the development and writing of this paper. Finally, we thank the University of Melbourne AI group for providing funding for the annotation work.

 This research was supported by the Commonwealth through an Australian Government Research Training Program Scholarship (\url{https://doi.org/10.82133/C42F-K220}) and conducted using the LIEF HPC-GPGPU Facility hosted at the University of Melbourne. This Facility was established with the assistance of LIEF Grant LE170100200. JHL was supported by the Australian Research Council under Grant LP210200917 and DP240101006. LF was supported by the Australian Research Council under Grant DE230100761.

\bibliography{custom}
\newpage

\appendix

\section{Implementation details}

\subsection{Dialogue segmentation implementation} \label{appendix:segment}
We segment each discussion into coherent topical units using an LLM-based boundary detector. Given the topic and prior context, the model proposes a list of segment intervals over utterance indices (please refer to Table \ref{tab:seg_prompt_schema} for the prompt), targeting segments of roughly 500--800 words and no more than 20 turns. To improve robustness, we run the segmenter \(p{=}5\) times with diversified decoding and convert each run into a set of proposed breakpoints (segment starts). We then perform \textbf{weighted majority voting} over candidate breakpoints: for each proposed start index \(b\), we add a vote of \(1.0\) to \(b\) and \(\,0.5\) to its immediate neighbors \((b{-}1, b{+}1)\), which makes the voting tolerant to small off-by-one differences across runs. Aggregating and normalizing votes yields a breakpoint confidence profile over utterances. Final breakpoints are selected as local peaks exceeding a threshold while enforcing minimum/maximum segment-length constraints, with a fallback to the highest-scoring breakpoint only when a cut is required to satisfy the maximum length. Finally, we merge adjacent segments when doing so reduces deviation from the desired word-count range (default 450--750 words) without exceeding an utterance cap, yielding a small set of well-formed segments per session.

\begin{table}[H]
\centering
\small
\resizebox{\columnwidth}{!}{
\begin{tabular}{p{0.20\linewidth} p{0.74\linewidth}}
\toprule
\textbf{Section} & \textbf{Prompt part (abridged)} \\ \midrule

\rowcolor[HTML]{EFEFEF}
Role \& task &
You are an expert dialogue analyst. Segment the interaction phase into coherent subtopic segments of \textbf{450--700 words} or \textbf{$\leq$20 speaker turns} each. \\

Inputs (template vars) &
Topic/Goal: \texttt{\{topic\}} \newline
Prior dialogue summary: \texttt{\{prior\_summary\}} \\

\rowcolor[HTML]{EFEFEF}
Instruction &
Identify subtopics and boundaries for \emph{all} utterances in the provided dialogue.\newline
Return \textbf{only} valid JSON matching the registered schema (no extra text).\newline
\texttt{segment\_index} starts from 0. \\

Output format &
\texttt{[\{segment\_index:int,\ utterances\_interval:[int,int],\ segment\_subtopic:str\},\ \ldots]} \\

\rowcolor[HTML]{EFEFEF}
Field definitions &
\texttt{segment\_index}: 0-based segment id.\newline
\texttt{utterances\_interval}: \texttt{[start\_idx, end\_idx]} over utterance indices.\newline
\texttt{segment\_subtopic}: short description of the segment's subtopic. \\

\bottomrule
\end{tabular}
}
\caption{Dialogue segmentation prompt and output schema. Curly-brace fields (e.g., \texttt{\{topic\}}) denote template variables filled at runtime.}
\label{tab:seg_prompt_schema}
\end{table}

\subsection{Segment selection for annotation}\label{sec:segment_selection}
To keep the human annotation workload feasible and affordable while still sampling informative portions of each episode, we select $k{=}4$ segments per conversation using a constrained, score-based procedure implemented in our task-generation script. For every candidate segment, we first filter the target utterances by marking moderator/audience turns as \textit{skipped}, and also skipping truncated or fragmentary utterances (e.g., $\leq$5 words without terminal punctuation, or $\leq$3 words), since these are most likely low in informativeness. We then estimate the total reading time of the task by summing (i) the segment utterances, (ii) the generated prior-context summary shown to the annotator, and (iii) a short window of immediately preceding dialogue (up to $K{=}5$ turns) that fits within a fixed reading-time budget; the prior-history window is truncated if adding more context would exceed the budget. 

Next, we compute a \emph{segment quality} vector that balances (a) task feasibility and (b) expected annotation value. Feasibility terms include deviation from a target number of non-skipped utterances (to avoid segments that are too sparse or too dense), deviation from the reading-time budget, and the segmenter’s confidence (mean boundary confidence). Expected value is approximated using the \textit{context benefit} of the chosen summary type: for the same segment, we compare GPT-5’s CIG/aspect scoring consistency under the selected context (e.g., \texttt{memory\_summary}) versus a \texttt{no\_summary} baseline, using the segment’s stored \texttt{summary\_scores}; segments where the summary yields a larger improvement are preferred because they are more sensitive to having an accurate prior-knowledge representation. Finally, we rank segments within each conversation using these features (via \texttt{rank\_segments}) and take the top $k{=}4$ ranked segments, while retaining metadata such as skip ratios, participant count, and reading-time breakdown for reporting and audit.

\subsection{Memory-Based Summarisation}\label{sec:memory_summarisation}

\begin{table}[h!]
\centering
\small
\resizebox{\columnwidth}{!}{
\begin{tabular}{p{0.20\linewidth} p{0.74\linewidth}}
\toprule
\textbf{Section} & \textbf{Prompt part (abridged)} \\ \midrule

\rowcolor[HTML]{EFEFEF}
Role \& task &
You are an expert dialogue analyst. Produce a coherent, highly readable summary of the prior context that is useful for interpreting the current segment. \\

Inputs (template vars) &
Topic: \texttt{\{topic\}} \newline
Prior context: \texttt{\{prior\_dialogue\}} \textit{or} \texttt{\{prior\_summary\}} \textit{or} \texttt{\{formatted\_memories\}} \newline
Current segment: \texttt{\{current\_dialogue\}} \\

\rowcolor[HTML]{EFEFEF}
Length \& style constraints &
Plain prose; no bullet points.\newline
Target length: \(\sim\)250 words (bounded range enforced by the prompt).\newline
Summary must begin with: \texttt{"The prior conversation..."} \\

Faithfulness constraint &
Use only information available in the provided prior context (no hallucination). \\

\rowcolor[HTML]{EFEFEF}
Output \& formatting &
Return \textbf{only} a JSON object (no extra text, no markdown):\newline
\texttt{\{"summary":"<two-paragraph summary>"\}} \\

Variants (prior-context representation) &
\textbf{Direct summary:} condition on \texttt{\{prior\_dialogue\}}.\newline
\textbf{Recursive summary:} update \texttt{\{prior\_summary\}} using \texttt{\{current\_dialogue\}}.\newline
\textbf{Theme-aware summary:} extract salient themes from \texttt{\{current\_dialogue\}}, then summarise relevant parts of \texttt{\{prior\_dialogue\}}.\newline
\textbf{Memory-based summary:} condition on retrieved memories \texttt{\{formatted\_memories\}}. \\

\bottomrule
\end{tabular}
}
\caption{Summarisation prompt template (abridged) and its four variants, which differ only in how the prior context is represented (full transcript, recursive prior summary, theme-aware decomposition, or retrieved semantic memories). Curly-brace fields (e.g., \texttt{\{topic\}}) denote template variables filled at runtime.}
\label{tab:summary_prompt_schema}
\end{table}

Scoring CIG requires a \emph{knowledge context} that reflects what has already been established in the discussion. Providing the full preceding transcript as context is often impractical: it is token-expensive for LLM-based scoring and cognitively burdensome for human annotators, who must judge informativeness relative to what the group already knows. While our memory module can retrieve semantically relevant prior claims, presenting the retrieved claim list directly to annotators can also be overwhelming and fragmentary, especially in long multi-party episodes. We therefore summarise the prior context into a short, readable prior-knowledge passage that (i) reduces prompt length for automated scoring and (ii) makes long episodes digestible for annotation without discarding the key claims and points of contention needed to judge Novelty, Relevance, Implication Scope, and overall CIG.

To develop and validate this summarisation procedure, we implemented multiple prior-context variants (direct transcript-based summary; prior-summary recursion; theme-aware decomposition; and a memory-based summary conditioned on retrieved semantic memories). We then ran a controlled comparison on one fully author-annotated seed episode (198 utterances): for each segment, we generated each context variant, asked GPT-5 to score overall CIG and the three aspects under that context, and compared these predictions against the author labels. Table~\ref{tab:summary_ablation} shows that weak-context baselines (e.g., \textit{no\_summary} and \textit{prior\_summary}) yield substantially higher error and lower agreement---most notably for \textit{info} and \textit{novo}---whereas memory-based summaries consistently improve alignment (lower MSE, higher $\kappa$) while remaining comparable to transcript-dependent summaries. This result motivates our use of retrieved-memory summaries as the default prior context for segment-level CIG scoring.

\begin{table}[t]
\footnotesize
\setlength{\tabcolsep}{4pt}
\renewcommand{\arraystretch}{1.15}
\resizebox{\columnwidth}{!}{
\begin{tabular}{p{0.4\columnwidth}  p{0.1\columnwidth} p{0.1\columnwidth} p{0.1\columnwidth}p{0.1\columnwidth}p{0.1\columnwidth}}
\toprule
\textbf{Context variant} & \textbf{Aspect} & \textbf{Mean} & \textbf{SD} & \textbf{MSE}$\downarrow$ & $\boldsymbol{\kappa}\uparrow$ \\
\midrule
\multirow{4}{*}{Human (reference)} 
& imsc & 2.292 & 1.179 & -- & -- \\
& info & 1.876 & 0.932 & -- & -- \\
& novo & 2.022 & 0.916 & -- & -- \\
& relv & 3.044 & 1.158 & -- & -- \\
\midrule
\multirow{4}{*}{No summary} 
& imsc & 2.449 & 1.094 & 0.916 & 0.649 \\
& info & 2.256 & 1.156 & 1.044 & 0.556 \\
& novo & 2.310 & 1.182 & 1.178 & 0.492 \\
& relv & 3.157 & 1.244 & 1.018 & 0.649 \\
\midrule
\multirow{4}{*}{Prior-summary only} 
& imsc & 2.464 & 1.130 & 0.952 & 0.647 \\
& info & 2.092 & 1.054 & 0.756 & 0.628 \\
& novo & 1.993 & 1.096 & 0.854 & 0.582 \\
& relv & 3.197 & 1.204 & 1.088 & 0.613 \\
\midrule
\multirow{4}{*}{Direct summary} 
& imsc & 2.408 & 1.100 & 0.906 & 0.654 \\
& info & 2.088 & 1.024 & 0.686 & 0.651 \\
& novo & 1.974 & 1.074 & 0.800 & 0.600 \\
& relv & 3.237 & 1.162 & 0.953 & 0.650 \\
\midrule
\multirow{4}{*}{Aspect-aware summary} 
& imsc & 2.398 & 1.108 & 0.850 & 0.677 \\
& info & 2.033 & 1.030 & 0.616 & 0.684 \\
& novo & 1.912 & 1.063 & 0.744 & 0.625 \\
& relv & 3.146 & 1.238 & 0.913 & 0.684 \\
\midrule
\multirow{4}{*}{Memory-based summary} 
& imsc & 2.343 & 1.084 & 0.847 & 0.671 \\
& info & 2.010 & 1.016 & 0.595 & 0.690 \\
& novo & 1.916 & 1.072 & 0.785 & 0.607 \\
& relv & 3.146 & 1.241 & 0.956 & 0.671 \\
\midrule
\multirow{4}{*}{Full transcript (LLM)} 
& imsc & 2.274 & 1.104 & 0.821 & 0.686 \\
& info & 1.985 & 1.025 & 0.548 & 0.716 \\
& novo & 1.912 & 1.046 & 0.730 & 0.624 \\
& relv & 3.102 & 1.228 & 0.971 & 0.660 \\
\bottomrule
\end{tabular}
}
\caption{
Summary-context ablation on one author-annotated seed session. ``Mean''/``SD'' report the rating distribution under each context variant; MSE and $\kappa$ measure agreement against the author ratings (human reference row has no agreement scores). Lower MSE and higher $\kappa$ indicate closer alignment.
}
\label{tab:summary_ablation}
\end{table}

\subsection{Memory extraction and consolidation}\label{sec:memory_modules}

To build a compact, queryable representation of what has been established so far in a multi-party discussion, we use a two-stage memory module: (i) \emph{atomic claim extraction} from each target utterance, and (ii) \emph{memory consolidation} that decides how each new claim should affect the evolving memory state. This design separates \emph{content decomposition} (turn $\rightarrow$ propositions) from \emph{state management} (propositions $\rightarrow$ persistent memory), allowing the downstream summarisation and CIG scoring components to operate over a stable set of self-contained claims rather than raw transcript text.

\begin{table*}[h]
\centering
\small
\setlength{\tabcolsep}{6pt}
\renewcommand{\arraystretch}{1.2}
\resizebox{\textwidth}{!}{
\begin{tabular}{@{}lll p{7cm}@{}}
\toprule
\textbf{NLI Relation} & \textbf{Action} & \textbf{Why} & \textbf{Example (A=new, B=existing)} \\
\midrule
equivalent & \textsc{None} & A $\Leftrightarrow$ B; duplicate, keep older & 
\textbf{A}: “I have a cat” vs \textbf{B}: “My pet is a cat” \\
forward\_entail & \textsc{Update} & A $\Rightarrow$ B; A refines/extends B (merge) & 
\textbf{A}: “I have a black cat.” entails \textbf{B}: “I have a cat.” \\
backward\_entail & \textsc{None} & B $\Rightarrow$ A; A is weaker/redundant & 
\textbf{A}: “I have a cat.” is entailed by \textbf{B}: “I have a black cat.” \\
contradiction & \textsc{Update} & A $\perp$ B; correct/replace B with A & 
\textbf{A}: “I have a black cat.” vs \textbf{B}: “I have a white cat.” \\
neutral & \textsc{Add} & No strong link; new topic/info & 
\textbf{A}: “Decision deadline is next Monday.” (no related B) \\
\bottomrule
\end{tabular}
}
\caption{Memory Consolidator rules: Mapping from NLI relation to memory action.}
\label{tab:nli-to-action-same}
\end{table*}

\paragraph{Atomic claim extraction.}
The extractor prompt (Table~\ref{tab:claim_extraction_prompt}) instructs the model to act as an \textbf{Atomic-Fact Extractor} and output a list of \emph{atomic}, \emph{self-contained}, and \emph{semantically distinct} propositions for a given target utterance. We require claims to be understandable without the original dialogue by resolving pronouns and deictic references using the provided context, but we constrain the extractor to \emph{extract only from the target utterance} (context is used only for disambiguation). To avoid superficial surface variation, the prompt explicitly discourages paraphrase duplicates and speech-act descriptions (e.g., ``asks/suggests''), and removes hedges and filler language unless a speech verb is needed for reported speech. The extractor returns only JSON in a fixed schema---a list of proposition objects with \texttt{speaker}, \texttt{target\_speaker}, \texttt{claim}, and \texttt{turn\_id} fields---and is capped at 30 claims per utterance to prioritize salience and keep the memory growth controlled.

\paragraph{Memory consolidation via speaker-aware NLI.}
Given the existing memory state and a set of newly extracted claims, the consolidator prompt (Table~\ref{tab:memory_consolidation_prompt}) updates memory using a deterministic, speaker-aware procedure grounded in bidirectional natural language inference (NLI). For each new claim $A$, we first retrieve candidate existing memories $B$ that are semantically similar (prioritizing same-speaker items), then classify the logical relationship between $A$ and $B$ in both directions ($A\Rightarrow B$ and $B\Rightarrow A$). The bidirectional NLI outcomes are mapped to a compact relation set (\textit{equivalent}, \textit{forward\_entail}, \textit{backward\_entail}, \textit{contradiction}, \textit{neutral}), and a strict target-selection ladder chooses \emph{exactly one} eligible $B$ (or none) to ensure consistent, auditable behavior.

\paragraph{ADD/UPDATE/NONE decisions and update semantics.}
The consolidator then maps each relation to one of three actions: \textsc{ADD} when a claim is novel (\textit{neutral}) or comes from a different speaker (to preserve multi-party disagreement), \textsc{NONE} when a claim is redundant (\textit{equivalent} or \textit{backward\_entail}), and \textsc{UPDATE} when a claim refines or corrects an existing one (\textit{forward\_entail} or \textit{contradiction}). Updates are defined explicitly: contradictions replace the target memory, while forward entailment merges $A$ and $B$ into a more specific proposition. The module outputs a JSON list of \texttt{memory\_updates} containing the action, the inferred logical relation, the source claim, and the selected target memory (or \texttt{null}), enabling downstream components to reconstruct the memory timeline and compute memory-dynamics features.

\subsection{Annotation interface design}\label{sec:annotation_interface}

\begin{figure*}[t]
    \centering
    \includegraphics[width=\linewidth]{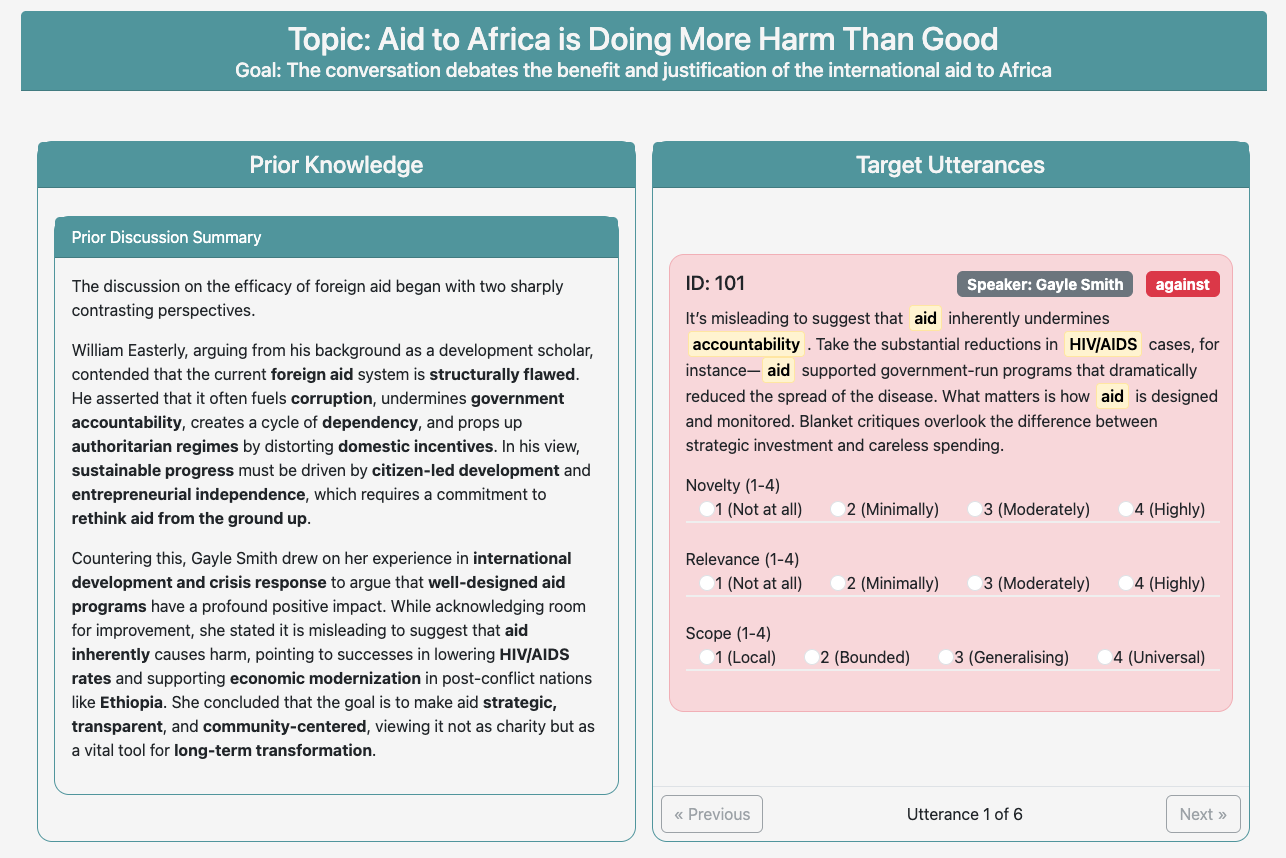}
    \caption{Annotation interface for segment-level CIG aspect rating. The left panel presents the prior-knowledge summary for the current segment, and the right panel displays target utterances with per-utterance ratings for Novelty, Relevance, and Implication Scope. The UI highlights overlapping keywords and supports click-to-scroll cross-referencing.}
    \label{fig:annotation_ui}
\end{figure*}

Figure~\ref{fig:annotation_ui} shows our web-based annotation interface, designed to make CIG judgments explicitly relative to shared prior context. The UI is organized into two synchronized panels. The left panel presents the \textit{Prior Knowledge} for the current segment as a compact prior-discussion summary with 3 preceding utterances, while the right panel presents the \textit{Target Utterances} one at a time, including the utterance ID, speaker name, and stance badge (e.g., \textit{for}/\textit{against}). For each utterance, annotators provide ordinal ratings for Novelty, Relevance, and Implication Scope using fixed four-level radio scales, and can navigate through utterances via \textit{Previous/Next} controls with a progress indicator (e.g., ``Utterance 1 of 6''). To reduce superficial scoring and encourage deliberate grounding in context, the right-hand rating panel is locked for the first 60 seconds whenever a new segment is loaded, forcing annotators to read the prior-knowledge summary before entering labels. In addition, the interface highlights keywords that appear in both the prior summary and the current utterance; clicking a highlighted keyword automatically scrolls the prior-knowledge panel to the corresponding mention, supporting rapid cross-referencing and helping annotators verify whether an utterance is truly novel or merely restating earlier claims.

\begin{table*}[h]
\centering
\small
\setlength{\tabcolsep}{6pt}
\renewcommand{\arraystretch}{1.15}
\resizebox{\textwidth}{!}{
\begin{tabular}{p{0.17\linewidth} p{0.08\linewidth} p{0.55\linewidth} p{0.18\linewidth}}
\toprule
\multicolumn{4}{p{\linewidth}}{\textbf{Context.} \textbf{Topic:} \textit{Gun Reduces Crime}.} \\
\multicolumn{4}{p{\linewidth}}{\textbf{Prior summary:} Pro speakers (Gary Kleck, Gayle Smith) argue that responsible firearm ownership deters criminals and enhances safety; opponent (R.\ Gil Kerlikowske) argues that widespread firearm availability increases risk to civilians and law enforcement and that tighter controls and prevention are more effective.} \\
\multicolumn{4}{p{\linewidth}}{\textbf{Immediate preceding turn (moderator):} \textit{``\ldots you were talking about British police being unarmed\ldots and yet Gary Kleck\ldots said why are police armed\ldots unless it is to deter assault\ldots can you take that on?''}} \\
\midrule
\textbf{Aspect} & \textbf{Level} & \textbf{Example utterance (given the context above)} & \textbf{Why this level} \\
\midrule

\multirow{4}{*}{Overall CIG} 
& 1 & R.\ Gil Kerlikowske: ``That's just wrong---guns don't reduce crime.'' 
& Pure assertion; adds no new content/mechanism. \\

& 2 & R.\ Gil Kerlikowske: ``To clarify, I'm talking about police carrying firearms, not civilian ownership.'' 
& Narrow clarification; limited informational advance. \\

& 3 & R.\ Gil Kerlikowske: ``If officers are armed, some attackers may back off, which could reduce assaults on police even if other crimes don't change.'' 
& Adds a concrete mechanism within the existing deterrence frame. \\

& 4 & R.\ Gil Kerlikowske: ``We should separate \emph{deterrence} from \emph{escalation}: arming police may deter some assaults but can also raise the lethality of encounters, so ``reduces crime'' depends on which outcomes we count.'' 
& Reframes the evaluation criterion; redirects the discussion. \\

\midrule
\multirow{4}{*}{Novelty} 
& 1 & Gary Kleck: ``Police carry guns to deter criminals---that's the point.'' 
& Restates a point already present in the prior summary. \\

& 2 & Gayle Smith: ``In some places, officers patrol without firearms.'' 
& Small factual addition; limited conceptual novelty. \\

& 3 & R.\ Gil Kerlikowske: ``Even the \emph{possibility} of a gun being taken from an officer changes how police approach routine encounters.'' 
& Introduces a specific explanatory detail not established in context. \\

& 4 & Gary Kleck: ``Deterrence is a \emph{belief} effect: what matters is offenders' perceived risk of armed resistance, not simply the raw number of guns.'' 
& Introduces a new conceptual lens/principle. \\

\midrule
\multirow{4}{*}{Relevance} 
& 1 & Gayle Smith: ``Seattle is a hard place to drive in.'' 
& Off-topic with no connection to the motion. \\

& 2 & R.\ Gil Kerlikowske: ``Public trust in police affects whether people cooperate with investigations.'' 
& Indirect link; requires bridging inference to connect to crime reduction via guns. \\

& 3 & R.\ Gil Kerlikowske: ``Officer assaults are related to public safety, but they are not the full question of whether guns reduce overall crime.'' 
& Clearly connected but peripheral to the core claim. \\

& 4 & Gary Kleck: ``If guns deter offenders, then wider lawful ownership can reduce crime---that's the motion we're debating.'' 
& Directly addresses the core topic/goal. \\

\midrule
\multirow{4}{*}{Implication Scope} 
& 1 & R.\ Gil Kerlikowske: ``Could you repeat the question?'' 
& Procedural/local to the interaction. \\

& 2 & Gayle Smith: ``I have a 2 years old boy.'' 
& Personal information; bounded to an individual case. \\

& 3 & R.\ Gil Kerlikowske: ``My neighborhood has been unsafe in the last 5 years, and I think it is not just that, but wider region.'' 
& Generalizes beyond a single case to a broader pattern. \\

& 4 & Gary Kleck: ``Public policy should preserve a general right of self-defense while minimizing harms to the broader community.'' 
& Principle-level claim with wide public applicability. \\

\bottomrule
\end{tabular}
}
\caption{
Exemplar utterances illustrating Levels 1--4 for Overall CIG and its three aspects under a fixed deliberation context. Speaker names are restricted to those appearing in the prior summary. Examples are illustrative and not drawn verbatim from the corpus.
}
\label{tab:cig_rubric_example}
\end{table*}

\begin{figure*}[h]
  \centering
  \includegraphics[width=0.7\linewidth]{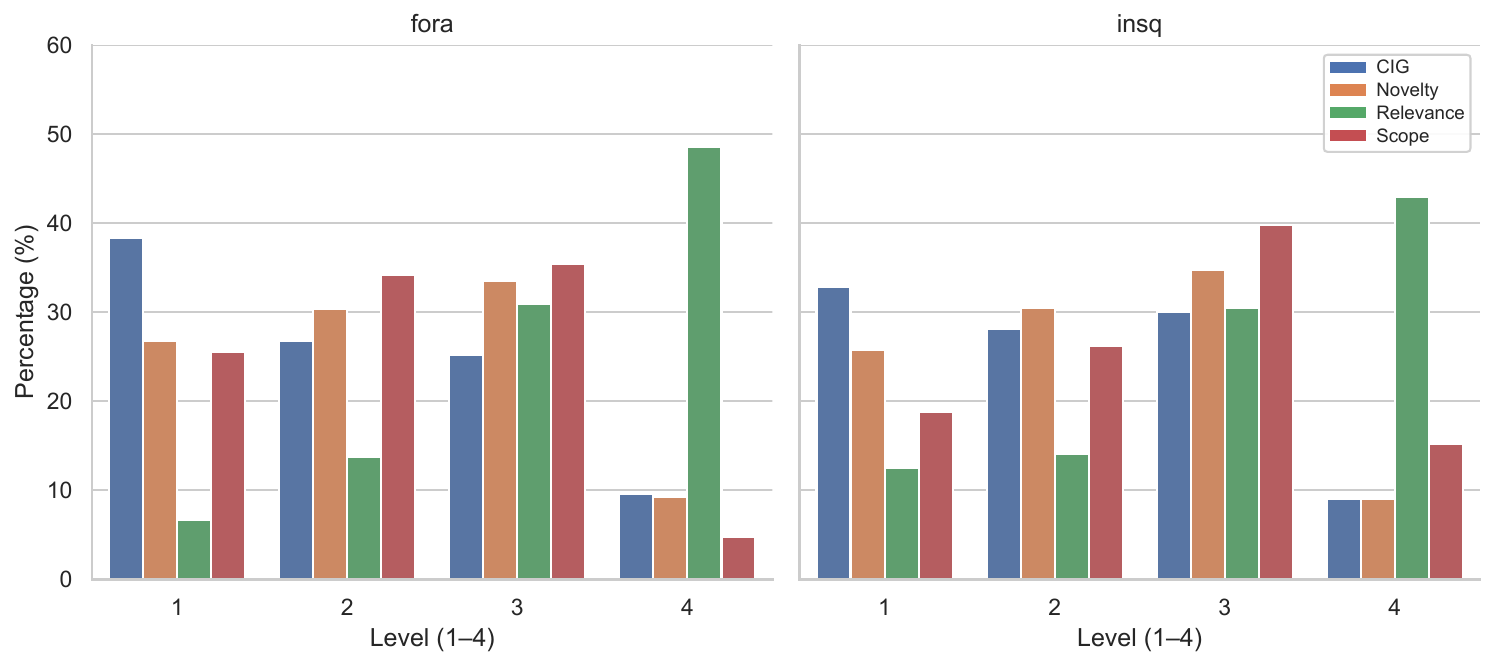}
  \caption{Normalized distributions of annotation levels (1–4) per aspect, and by corpus.}
  \label{fig:label-distributions}
\end{figure*}

\begin{table*}[t]
\centering
\small
\resizebox{\textwidth}{!}{
\begin{tabular}{p{0.20\linewidth} p{0.74\linewidth}}
\toprule
\textbf{Section} & \textbf{Prompt part (abridged)} \\ \midrule

\rowcolor[HTML]{EFEFEF}
Role &
You are an \textbf{Atomic-Fact Extractor}. \\

Goal &
Given a conversational context and a target utterance, extract a list of \textbf{atomic claims}. \\

\rowcolor[HTML]{EFEFEF}
Core principles &
\textbf{Self-Contained:} each claim is understandable without the original dialogue.\newline
\textbf{Atomic:} each claim is the smallest proposition that can be independently true or false.\newline
\textbf{Semantically Distinct:} do not output paraphrase duplicates; each claim must be unique. \\

Extraction rules &
(1) Extract explicit claims literally stated in the target utterance.\newline
(2) Extract salient implicit claims a listener would confidently infer.\newline
(3) Focus on content, not speech acts (avoid describing asking/suggesting). \\

\rowcolor[HTML]{EFEFEF}
What to avoid &
Do not describe the act of speaking (avoid ``Asks...'', ``States...'', ``Suggests...'').\newline
Do not include hedge/filler (e.g., ``I think'', ``maybe'', ``kind of'').\newline
Exception: a speech verb is allowed only for reported speech (e.g., ``Stephen said that ...''). \\

Output schema &
Return \textbf{only} valid JSON:\newline
\texttt{\{"memories":[\{"speaker":"...", "target\_speaker":"...", "claim":"...", "turn\_id":"..."\}, ...]\}}\newline
If none: \texttt{\{"memories":[]\}} \\

\rowcolor[HTML]{EFEFEF}
Quality checks &
Use the \textbf{Context} only to resolve ambiguity (pronouns/deictics); extract only from the \textbf{Target utterance}.\newline
One JSON object per proposition; do not extract more than 30 claims (keep the most important).\newline
If speaker appears as ``Name (role)'', keep only the full name.\newline
Return JSON only (no markdown, no extra text). \\

Example (from prompt) &
\textbf{Context:}\newline
1.\ Speaker 1: I hope my kids own guns.\newline
2.\ Speaker 2: I am thinking the opposite.\newline
\textbf{Target utterance:}\newline
3.\ Speaker 3: When I look at the statistics about how that adds to the risk of suicide, the risk of being misused, the risk of it being stolen, used in a domestic quarrel, I think it's just too much of a risk.\newline
\textbf{Output:}\newline
\texttt{\{"memories":[}\newline
\texttt{\ \{"speaker":"Speaker 3","target\_speaker":"Everyone","claim":"Having a gun increases the risk of suicide.","turn\_id":"3"\},}\newline
\texttt{\ \{"speaker":"Speaker 3","target\_speaker":"Everyone","claim":"Having a gun increases the risk of misuse.","turn\_id":"3"\},}\newline
\texttt{\ \{"speaker":"Speaker 3","target\_speaker":"Everyone","claim":"Having a gun increases the risk of theft.","turn\_id":"3"\},}\newline
\texttt{\ \{"speaker":"Speaker 3","target\_speaker":"Everyone","claim":"Having a gun increases the risk of use in domestic quarrels.","turn\_id":"3"\},}\newline
\texttt{\ \{"speaker":"Speaker 3","target\_speaker":"Everyone","claim":"Owning a gun is too risky.","turn\_id":"3"\}}\newline
\texttt{]\}} \\

\rowcolor[HTML]{EFEFEF}
Other examples &
Example A, B, C, E, F, \ldots \\

\bottomrule
\end{tabular}
}
\caption{
\textbf{Abridged} single-task prompt for multi-party atomic claim extraction. The table summarizes the role, constraints, output schema, and one representative in-context example; remaining examples are omitted for space.
}
\label{tab:claim_extraction_prompt}
\end{table*}

\begin{table*}[h]
\centering
\small
\resizebox{\linewidth}{!}{
\begin{tabular}{p{0.20\linewidth} p{0.74\linewidth}}
\toprule
\textbf{Section} & \textbf{Prompt part (abridged)} \\ \midrule

\rowcolor[HTML]{EFEFEF}
Role &
You are a \textbf{Multi-Party Memory Consolidator}. For each newly extracted claim, decide whether to \textbf{ADD}, \textbf{UPDATE}, or \textbf{NONE} using NLI. \\

Input &
\texttt{existing\_memories}: JSON array of stored proposition objects.\newline
\texttt{newly\_extracted\_claims}: JSON array of new proposition objects. \\

\rowcolor[HTML]{EFEFEF}
Core procedure &
(1) Retrieve candidates: search for the most semantically relevant existing memory, prioritizing same-speaker items.\newline
(2) NLI: classify relation between $A$ (new) and $B$ (existing) in both directions ($A\Rightarrow B$, $B\Rightarrow A$).\newline
(3) Decide action: apply speaker-aware decision rules.\newline
(4) Emit one JSON update object for $A$. \\

NLI relation set &
Map bidirectional NLI to one of:\newline
\textit{equivalent} ($A\Leftrightarrow B$), \textit{forward\_entail} ($A\Rightarrow B$ only), \textit{backward\_entail} ($B\Rightarrow A$ only), \textit{contradiction}, \textit{neutral}. \\

\rowcolor[HTML]{EFEFEF}
Target selection (priority) &
Choose \emph{exactly one} target $B$ (or none) by the following ladder:\newline
1) same speaker \& \textit{equivalent}\newline
2) same speaker \& \textit{backward\_entail}\newline
3) same speaker \& (\textit{contradiction} or \textit{forward\_entail})\newline
4) different speaker \& any non-neutral relation\newline
Else: no eligible $B$ $\rightarrow$ treat as neutral (ADD, target=null).\newline
Ties within a rung: pick the highest confidence (or highest similarity). \\

Action mapping &
Same speaker:\newline
\textit{equivalent}, \textit{backward\_entail} $\rightarrow$ \textsc{NONE};\quad
\textit{forward\_entail}, \textit{contradiction} $\rightarrow$ \textsc{UPDATE};\quad
\textit{neutral} $\rightarrow$ \textsc{ADD}.\newline
Different speaker:\quad always \textsc{ADD}. \\

\rowcolor[HTML]{EFEFEF}
UPDATE semantics &
If \textit{contradiction}: replace $B$ with $A$ (source is $A$).\newline
If \textit{forward\_entail}: merge $A$ with $B$ into a more specific claim (source is merged claim). \\

Output schema &
\texttt{\{"memory\_updates":[\{"action":"ADD|UPDATE|NONE", "logical\_relation":"...", "source":\{...\}, "target":null|\{id,...\}\}]\}}\newline
Return \texttt{\{"memory\_updates":[]\}} if nothing changes. \\

\rowcolor[HTML]{EFEFEF}
Quality checks &
Rewrite claims to be context-independent; resolve pronouns; remove hedges/filler.\newline
\texttt{target\_speaker} denotes a person/group (not an object).\newline
\texttt{target} may only include: \texttt{id, speaker, target\_speaker, claim, turn\_id}. \\

Example (from prompt) &
\textbf{Existing memory:}\newline
\texttt{\{ "id":"mem\_012", "speaker":"Sam", "target\_speaker":"Everyone",}\newline
\texttt{\ \ "claim":"The financial cost of death penalty appeals is very high.", "turn\_id":"8"\}}\newline
\textbf{New claim $A$:}\newline
\texttt{\{ "speaker":"Sam", "target\_speaker":"Everyone",}\newline
\texttt{\ \ "claim":"The financial cost of death penalty appeals exceeds the cost of life imprisonment.", "turn\_id":"14"\}}\newline
\textbf{Output update:}\newline
\texttt{\{ "action":"UPDATE", "logical\_relation":"forward\_entail",}\newline
\texttt{\ \ "source":\{...\ turn 14 ...\}, "target":\{...\ mem\_012 ...\}\}} \\

\bottomrule
\end{tabular}
}
\caption{Abridged prompt for multi-party memory consolidation, including the NLI relation set, deterministic target-selection ladder, and action mapping that produces \textsc{ADD}/\textsc{UPDATE}/\textsc{NONE} updates. One illustrative example is shown; additional examples in the original prompt are omitted for brevity.}
\label{tab:memory_consolidation_prompt}
\end{table*}

\begin{table*}[h]
\centering
\small
\resizebox{\textwidth}{!}{
\begin{tabular}{p{0.20\textwidth} p{0.74\textwidth}}
\toprule
\textbf{Section} & \textbf{Prompt part (abridged)} \\ \midrule

\rowcolor[HTML]{EFEFEF}
Role \& task &
You are an expert dialogue analyst acting from the perspective of a community audience. Rate each \textbf{TARGET} utterance on \textbf{Informativeness} (1--4). \\

Input (template vars) &
Topic/Goal: \texttt{\{topic\}}\newline
Prior knowledge + preceding dialogue: \texttt{\{context\}}\newline
Target utterances: \texttt{\{target\}} (evaluate indices \texttt{\{start\}}--\texttt{\{end\}}, total \texttt{\{total\}}). \\

\rowcolor[HTML]{EFEFEF}
Baseline assumption &
Treat the \texttt{Prior Knowledge} section and preceding dialogue as the \textbf{Shared Knowledge Baseline}; repeating/paraphrasing baseline content $\rightarrow$ low scores. \\

Dimension definition &
\textbf{Conversational Information Gain (Informativeness)}: how much the utterance advances shared understanding or progress on the topic, given baseline.\newline
Scale anchors: 1=no gain; 2=minimal; 3=incremental; 4=insightful. \\

\rowcolor[HTML]{EFEFEF}
Output \& schema &
Return \textbf{only} valid JSON (no extra text):\newline
\texttt{[\{"utterance\_index":int,\ "informativeness":int,\ "context\_type":INFO\},\ \ldots]} \\
& \textit{(\texttt{context\_type} is hard-coded for identification.)} \\

\rowcolor[HTML]{EFEFEF}
Example (from prompt) &
\textbf{Topic:} Gun Reduces Crime.\newline
\textbf{Prior knowledge:} brief debate summary (Kleck/Smith vs Kerlikowske).\newline
\textbf{Target utterances:} indexed 1--4.\newline
\textbf{Illustrative output:}\newline
\texttt{[\{"utterance\_index":1, "informativeness":3\},\ \{"utterance\_index":2, "informativeness":1\},\ \{"utterance\_index":3, "informativeness":1\},\ \{"utterance\_index":4, "informativeness":4\}]} \\

\bottomrule
\end{tabular}
}
\caption{
\textbf{Abridged} info-only rating prompt used to score segment utterances on Informativeness (CIG).
For space, we \emph{shorten} (i) the system preamble and reminder lines, (ii) the full prose definitions for the 1--4 scale, and (iii) the example's prior-knowledge summary and dialogue excerpt; the table preserves the exact \texttt{\{topic\}/\{context\}/\{target\}} inputs, rating dimensions, and the required JSON output schema.
}
\label{tab:rating_prompt_info}
\end{table*}

\begin{table*}[h]
\centering
\small
\begin{tabular}{p{0.20\linewidth} p{0.74\linewidth}}
\toprule
\textbf{Section} & \textbf{Prompt part (abridged)} \\ \midrule

\rowcolor[HTML]{EFEFEF}
Role \& task &
You are an expert dialogue analyst acting from the perspective of a community audience. Rate each \textbf{TARGET} utterance on \textbf{Novelty}, \textbf{Relevance}, and \textbf{Implication Scope} (each 1--4). \\

Input (template vars) &
Topic/Goal: \texttt{\{topic\}}\newline
Prior knowledge + preceding dialogue: \texttt{\{context\}}\newline
Target utterances: \texttt{\{target\}} (evaluate indices \texttt{\{start\}}--\texttt{\{end\}}, total \texttt{\{total\}}). \\

\rowcolor[HTML]{EFEFEF}
Baseline assumption &
Treat the \texttt{Prior Knowledge} section and preceding dialogue as the \textbf{Shared Knowledge Baseline}; repetitions $\rightarrow$ low \textbf{Novelty}. \\

Dimension definitions &
\textbf{Novelty}: newness relative to baseline.\newline
\textbf{Relevance}: substantive connection to topic/goal.\newline
\textbf{Implication Scope}: scope of intended reach/generality. \\

\rowcolor[HTML]{EFEFEF}
Output \& schema &
Return \textbf{only} valid JSON (no extra text):\newline
\texttt{[\{"utterance\_index":int,\ "novelty":int,\ "relevance":int,\ "implication\_scope":int,\ "context\_type":MIX\},\ \ldots]} \\
& \textit{(\texttt{context\_type} is hard-coded for identification.)} \\

\rowcolor[HTML]{EFEFEF}
Example (from prompt) &
\textbf{Topic:} Gun Reduces Crime.\newline
\textbf{Prior knowledge:} brief debate summary (Kleck/Smith vs Kerlikowske).\newline
\textbf{Target utterances:} indexed 1--4.\newline
\textbf{Illustrative output:}\newline
\texttt{[\{"utterance\_index":1,"novelty":3,"relevance":3,"implication\_scope":3\},}\newline
\texttt{\ \{"utterance\_index":2,"novelty":1,"relevance":4,"implication\_scope":4\},}\newline
\texttt{\ \{"utterance\_index":3,"novelty":3,"relevance":1,"implication\_scope":1\},}\newline
\texttt{\ \{"utterance\_index":4,"novelty":4,"relevance":3,"implication\_scope":3\}]} \\

\bottomrule
\end{tabular}
\caption{
\textbf{Abridged} mixed-aspects rating prompt used to score segment utterances on Novelty, Relevance, and Implication Scope.
For space, we \emph{shorten} (i) the system preamble and reminder lines, (ii) the full prose definitions and all anchor examples for each 1--4 scale, and (iii) the example's prior-knowledge summary and dialogue excerpt; the table preserves the exact \texttt{\{topic\}/\{context\}/\{target\}} inputs, the rated dimensions, and the required JSON output schema.
}
\label{tab:rating_prompt_mix}
\end{table*}

\begin{table*}[h]
\centering
\small
\resizebox{\textwidth}{!}{
\begin{tabular}{p{0.20\linewidth} p{0.74\linewidth}}
\toprule
\textbf{Section} & \textbf{Prompt part (abridged)} \\ \midrule

\rowcolor[HTML]{EFEFEF}
Role \& task &
You are an expert dialogue analyst acting from the perspective of a community audience. Rate each \textbf{TARGET claim} on \textbf{Informativeness (CIG)}, \textbf{Novelty}, \textbf{Relevance}, and \textbf{Implication Scope} (each 1--4). \\

Input (template vars) &
Topic/Goal: \texttt{\{topic\}}\newline
(Optional) Dialogue context: \texttt{\{dialogue\_context\}}\newline
Existing memories (shared baseline): \texttt{\{existing\_memories\}}\newline
Target claims: \texttt{\{claims\}} \\

\rowcolor[HTML]{EFEFEF}
Baseline assumption &
Treat \texttt{Existing memories} as the \textbf{Shared Knowledge Baseline}; repetitions/paraphrases of the baseline $\rightarrow$ low \textbf{Novelty} (and typically low \textbf{Informativeness}). \\

Dimension definitions &
\textbf{Informativeness (CIG)}: forward-looking impact on shared understanding/progress.\newline
\textbf{Novelty}: newness relative to baseline/context only.\newline
\textbf{Relevance}: substantive connection to the topic/goal (procedural/meta $\rightarrow$ low).\newline
\textbf{Implication Scope}: intended reach/generality of the claim (local $\rightarrow$ universal). \\

\rowcolor[HTML]{EFEFEF}
Output \& schema &
Return \textbf{only} valid JSON (no extra text):\newline
\texttt{[\{"id":int,\ "informativeness":int,\ "novelty":int,\ "relevance":int,\ "implication\_scope":int\},\ \ldots]} \\

\rowcolor[HTML]{EFEFEF}
Example (from prompt) &
\textbf{Topic:} Should our state retain the death penalty?\newline
\textbf{Existing memories:} e.g., ``provides justice for victims' families''; ``appeals are very costly''.\newline
\textbf{Target claims:} e.g., ``risk of executing an innocent person''; ``appeals cost exceeds life imprisonment''; plus off-topic claims.\newline
\textbf{Illustrative output:}\newline
\texttt{[\{"id":13, "informativeness":4, "novelty":4, "relevance":4, "implication\_scope":4\},}
\texttt{\ \{"id":14, "informativeness":3, "novelty":3, "relevance":4, "implication\_scope":4\},\ \ldots]} \\

\bottomrule
\end{tabular}
}
\caption{
\textbf{Abridged} claim-level rating prompt used to score extracted claims on Informativeness (CIG), Novelty, Relevance, and Implication Scope given an \texttt{existing\_memories} baseline.
For space, we \emph{shorten} (i) the system preamble and reminder lines, (ii) the full prose definitions and all anchor examples for each 1--4 scale, and (iii) the example's dialogue context and claim list; the table preserves the exact template inputs (\texttt{\{topic\}}, \texttt{\{dialogue\_context\}}, \texttt{\{existing\_memories\}}, \texttt{\{claims\}}), the rated dimensions, and the required JSON output schema.
}
\label{tab:claim_rating_prompt_full}
\end{table*}

\section{Informativeness-related proxy definitions}
\label{appendix:proxy_equations}

This appendix defines the informativeness-related proxy features used in our correlation analysis
(Table~\ref{tab:feat_corr_human}). To avoid equation overflow in ACL two-column format, we use compact symbols
in math and refer to implementation variable names (e.g., \texttt{token\_count}) in text.

\subsection{Length and volume}
\label{appendix:proxy_length}

\paragraph{Length (tokens).}
Let $T_i$ be the spaCy token sequence for utterance $i$ (spaces excluded; punctuation retained). Utterance length
(\texttt{token\_count}) is:
\begin{equation}
n^{\text{tok}}_i = |T_i|.
\end{equation}

\paragraph{Content-word volume.}
Let $C_i$ be the multiset of \emph{content} lemmas in utterance $i$, obtained by filtering spaCy tokens to
alphabetic tokens that are not stopwords, numerals, or punctuation, and then lemmatising (fallback to
lowercased surface form when lemma is unavailable). Content-token volume (\texttt{content\_token\_count}) is:
\begin{equation}
n^{\text{cont}}_i = |C_i|.
\end{equation}

\subsection{TF--IDF and lexical specificity (episode-level)}
\label{appendix:proxy_tfidf}

\paragraph{TF--IDF (episode-level).}
We fit a \texttt{TfidfVectorizer} \emph{per episode} (utterance-as-document), using the content-lemma token list
as the analyzer. For utterance $i$, let $W_i=\{w_{ij}\}$ denote its nonzero TF--IDF weights. We compute:
\begin{equation}
\begin{aligned}
s^{\text{tfidf}}_i &= \sum_{j} w_{ij}, \\
m^{\text{tfidf}}_i &= \max_{j} w_{ij}, \\
\mu^{\text{tfidf}}_i &= \frac{1}{|W_i|}\sum_{j} w_{ij}.
\end{aligned}
\end{equation}
These correspond to \texttt{tfidf\_sum}, \texttt{tfidf\_max}, and \texttt{tfidf\_mean}. Implementation:
\texttt{norm=None}, \texttt{min\_df=2}, \texttt{max\_df=0.95}.

\paragraph{Lexical specificity (IDF).}
Let $\texttt{idf}_j$ be the episode-level IDF from the fitted vectorizer, and let $J_i$ be indices of the terms
present in utterance $i$ (i.e., nonzero TF--IDF entries). We compute:
\begin{equation}
\begin{aligned}
\mu^{\text{idf}}_i &= \frac{1}{|J_i|}\sum_{j\in J_i} \texttt{idf}_j, \\
\tilde{\mu}^{\text{idf}}_i &= \operatorname{median}\!\left(\{\texttt{idf}_j\}_{j\in J_i}\right).
\end{aligned}
\end{equation}
These correspond to \texttt{specificity\_mean\_idf} and \texttt{specificity\_median\_idf}. If $J_i=\emptyset$,
we set the specificity values to $0$.

\subsection{Lexical and entity novelty (cumulative)}
\label{appendix:proxy_novelty}

\paragraph{Word novelty (cumulative).}
Let $V_{<i}$ be the set of content lemmas observed earlier in the same episode, and let $L_i$ be the set of
content lemmas in utterance $i$ (derived from $C_i$). Define $N_i = L_i \setminus V_{<i}$ as the set of novel
lemmas introduced at utterance $i$. We compute:
\begin{equation}
\begin{aligned}
n^{\text{new}}_i &= |N_i|, \\
\rho^{\text{new}}_i &= \frac{n^{\text{new}}_i}{\max(1, n^{\text{cont}}_i)}.
\end{aligned}
\end{equation}
These correspond to \texttt{novel\_word\_count} and \texttt{novel\_word\_density}. We also track cumulative
debug counters such as \texttt{seen\_vocab\_size\_so\_far}$=|V_{<i}|$.

\paragraph{Entity novelty (cumulative).}
Let $E_i$ be the multiset of named-entity surface forms in utterance $i$ (lowercased), $U_i=\operatorname{set}(E_i)$
the set of unique entity forms in the utterance, and $S_{<i}$ the set of entity forms observed earlier in the
episode. We compute:
\begin{equation}
\begin{aligned}
n^{\text{ent}}_i &= |E_i|, \\
n^{\text{ent-new}}_i &= |U_i \setminus S_{<i}|, \\
\rho^{\text{ent-new}}_i &= \frac{n^{\text{ent-new}}_i}{\max(1, |U_i|)}.
\end{aligned}
\end{equation}
These correspond to \texttt{entity\_count}, \texttt{novel\_entity\_count}, and \texttt{novel\_entity\_ratio}.
We use spaCy NER labels:
\texttt{PERSON, ORG, GPE, LOC, NORP, EVENT, WORK\_OF\_ART, LAW, PRODUCT, FAC, LANGUAGE}.
We additionally track token-normalized entity novelty:
\begin{equation}
\rho^{\text{ent-new-tok}}_i
= \frac{n^{\text{ent-new}}_i}{\max(1, n^{\text{tok}}_i)},
\end{equation}
corresponding to \texttt{novel\_entity\_density\_token}.

\subsection{LM surprisal and predictability}
\label{appendix:proxy_surprisal}

\paragraph{LM surprisal / entropy.}
For a causal language model, token surprisal (bits) is $s_t=-\log_2 p(w_t\mid w_{<t})$. Let $T$ be the number of
evaluated tokens (excluding special tokens; determined via \texttt{attention\_mask} and \texttt{start\_idx}). We
compute average and summed surprisal:
\begin{equation}
\begin{aligned}
\bar{s}_i &= \frac{1}{T}\sum_{t=1}^{T} s_t, \\
S_i &= \sum_{t=1}^{T} s_t = T\cdot \bar{s}_i.
\end{aligned}
\end{equation}
These correspond to \texttt{sent\_avg\_h} and \texttt{sum\_h}. We also compute a length-normalized variant:
\begin{equation}
\bar{s}^{\,\text{norm}}_i = \frac{\bar{s}_i}{\bar{h}(T)},
\end{equation}
where $\bar{h}(T)$ is the mean \texttt{sent\_avg\_h} among utterances of length $T$
(\texttt{norm\_sent\_avg\_h}). We used Llama-3.2-3B as the causal model.

\paragraph{Top-quartile token predictability.}
The pipeline also stores per-token log-probabilities $\ell_t=\log_2 p(w_t\mid w_{<t})$ in \texttt{tokens\_h}
(negative values). Let $\mathcal{Q}_i$ be the indices of the top 25\% of tokens in utterance $i$ ranked by
$\ell_t$ (i.e., the most predictable tokens). We compute:
\begin{equation}
q_i = \frac{1}{|\mathcal{Q}_i|}\sum_{t\in\mathcal{Q}_i} \ell_t,
\end{equation}
corresponding to \texttt{top\_quatile\_avg\_ent}. Note that despite the variable name, this aggregates
\emph{log-probabilities} rather than surprisals; higher values (less negative) indicate more predictable tokens.

\subsection{Memory dynamics}
\label{appendix:proxy_memory}

\paragraph{Memory dynamics (counts).}
Let $\mathcal{A}_i$ be the set of memory-update actions aligned to utterance $i$ (from segment-level
\texttt{memory\_actions}). We define:
\begin{equation}
\begin{aligned}
n^{\text{claim}}_i &= |\mathcal{A}_i|, \\
\Delta^{\text{mem}}_i &= \sum_{a\in\mathcal{A}_i} \mathbb{I}\!\left[\texttt{event}(a)\neq \texttt{NONE}\right].
\end{aligned}
\end{equation}
These correspond to \texttt{claim\_count} and \texttt{mem\_delta}.

\paragraph{Aspect-gated memory changes.}
Let $\hat{y}_{i,d}\in\{1,2,3,4\}$ be the predicted aspect rating for utterance $i$ and aspect $d$. For
$d\in\{\texttt{info, novo, relv, imsc}\}$, we gate memory changes by requiring $\hat{y}_{i,d}>2$:
\begin{equation}
\Delta^{\text{mem}}_{i,d}
=
\sum_{a\in\mathcal{A}_i}
\mathbb{I}\!\left[\texttt{event}(a)\neq \texttt{NONE}\right]\,
\mathbb{I}\!\left[\hat{y}_{i,d}>2\right].
\end{equation}
The triad-gated variant additionally requires \texttt{novo}, \texttt{relv}, and \texttt{imsc} all exceed the
midpoint:
\begin{equation}
\Delta^{\text{mem}}_{i,\text{triad}}
=
\sum_{a\in\mathcal{A}_i}
\mathbb{I}\!\left[\texttt{event}(a)\neq \texttt{NONE}\right]
\prod_{d\in D}\mathbb{I}\!\left[\hat{y}_{i,d}>2\right],
\end{equation}
where $D=\{\texttt{novo},\texttt{relv},\texttt{imsc}\}$. These implement the Relv$^{-}$-style features reported
in Table~\ref{tab:feat_corr_human} by counting memory changes only when the corresponding aspect is rated above
the midpoint. In the released code, gating labels are sourced from stored per-utterance aspect predictions
(e.g., \texttt{claim\_predictions["gpt-5"]}) and are only available for utterances included in rated segments
(via \texttt{mem\_rating\_used}).

\begin{table*}[h]
\centering
\small
\resizebox{\textwidth}{!}{
\begin{tabular}{llccccc|ccccc}
\toprule
& & \multicolumn{5}{c}{\textbf{Fora}} & \multicolumn{5}{c}{\textbf{Insq}} \\
\cmidrule(lr){3-7} \cmidrule(lr){8-12}
\textbf{Model} & \textbf{Context} & \textbf{CIG} & \textbf{Nov.} & \textbf{Rel.} & \textbf{Scope} & \textbf{Mean} & \textbf{CIG} & \textbf{Nov.} & \textbf{Rel.} & \textbf{Scope} & \textbf{Mean} \\
\midrule
GPT-5 & Full & 0.512 & 0.556 & 0.429 & 0.483 & 0.495 & \textbf{0.411} & 0.577 & 0.444 & 0.541 & \textbf{0.493} \\
 & Memory & 0.531 & 0.574 & 0.419 & 0.496 & 0.505 & 0.459 & 0.567 & \textbf{0.427} & 0.526 & 0.495 \\
 & No Knowledge & 0.594 & 0.700 & 0.422 & 0.488 & 0.551 & 0.470 & 0.700 & 0.436 & 0.539 & 0.536 \\
 & Short Prior & 0.572 & 0.620 & 0.417 & 0.495 & 0.526 & 0.459 & 0.637 & 0.440 & 0.536 & 0.518 \\
 & Summary & 0.520 & 0.556 & \textbf{0.414} & 0.479 & \textbf{0.492} & 0.457 & 0.587 & 0.452 & 0.529 & 0.506 \\
\midrule
GPT-5-Mini & Full & 0.551 & 0.574 & 0.456 & \textbf{0.438} & 0.505 & 0.487 & 0.726 & 0.529 & 0.484 & 0.557 \\
 & Memory & 0.591 & 0.557 & 0.467 & 0.453 & 0.517 & 0.501 & 0.697 & 0.535 & 0.467 & 0.550 \\
 & No Knowledge & 0.646 & 0.542 & 0.448 & 0.463 & 0.524 & 0.557 & 0.599 & 0.520 & 0.472 & 0.537 \\
 & Short Prior & 0.622 & 0.539 & 0.460 & 0.455 & 0.519 & 0.577 & 0.530 & 0.551 & 0.473 & 0.533 \\
 & Summary & 0.569 & 0.617 & 0.475 & 0.472 & 0.533 & 0.546 & 0.806 & 0.547 & 0.489 & 0.597 \\
\midrule
Gemini-2.5-Pro & Full & 0.527 & 0.609 & 0.549 & 0.566 & 0.563 & 0.471 & 0.601 & 0.610 & 0.518 & 0.550 \\
 & Memory & 0.520 & 0.531 & 0.547 & 0.581 & 0.545 & 0.480 & 0.519 & 0.551 & 0.501 & 0.513 \\
 & No Knowledge & 0.535 & 0.531 & 0.526 & 0.568 & 0.540 & 0.479 & \textbf{0.476} & 0.566 & 0.513 & 0.509 \\
 & Short Prior & 0.518 & 0.536 & 0.562 & 0.555 & 0.543 & 0.484 & 0.506 & 0.556 & 0.523 & 0.517 \\
 & Summary & \textbf{0.505} & 0.571 & 0.547 & 0.578 & 0.550 & 0.431 & 0.580 & 0.578 & 0.512 & 0.525 \\
\midrule
Gemini-2.5-Flash & Full & 0.563 & 0.616 & 0.622 & 0.474 & 0.569 & 0.543 & 0.653 & 0.615 & 0.496 & 0.577 \\
 & Memory & 0.560 & 0.543 & 0.575 & 0.476 & 0.539 & 0.554 & 0.600 & 0.502 & 0.467 & 0.531 \\
 & No Knowledge & 0.654 & 0.632 & 0.661 & 0.492 & 0.610 & 0.677 & 0.621 & 0.648 & 0.465 & 0.603 \\
 & Short Prior & 0.588 & 0.570 & 0.533 & 0.471 & 0.540 & 0.689 & 0.652 & 0.710 & \textbf{0.453} & 0.626 \\
 & Summary & 0.589 & 0.585 & 0.579 & 0.452 & 0.551 & 0.611 & 0.628 & 0.579 & 0.458 & 0.569 \\
\midrule
Qwen3-4B & Full & 0.622 & 0.571 & 0.467 & 0.593 & 0.563 & 0.597 & 0.656 & 0.589 & 0.597 & 0.610 \\
 & Memory & 0.643 & 0.523 & 0.425 & 0.599 & 0.547 & 0.726 & 0.617 & 0.509 & 0.569 & 0.605 \\
 & No Knowledge & 0.619 & 0.723 & 0.469 & 0.655 & 0.617 & 0.536 & 0.812 & 0.569 & 0.626 & 0.636 \\
 & Short Prior & 0.535 & 0.520 & 0.462 & 0.575 & 0.523 & 0.601 & 0.673 & 0.494 & 0.586 & 0.589 \\
 & Summary & 0.579 & \textbf{0.519} & 0.472 & 0.588 & 0.539 & 0.658 & 0.618 & 0.528 & 0.593 & 0.599 \\
\bottomrule
\end{tabular}
}
\caption{MAE results for different models, contexts, and aspects across FORA and INSQ corpora. The last column in each corpus group shows the mean MAE across the four aspects. The lowest MAE in each column is highlighted in bold.}
\label{tab:mae_results}
\end{table*}

\begin{table}[H]
\centering
\footnotesize
\setlength{\tabcolsep}{6pt}
\renewcommand{\arraystretch}{1.15}
\begin{tabular}{l c}
\toprule
\textbf{Feature} & \textbf{$|r|$ w/ \texttt{CIG}} \\
\midrule
Memory changes (Novo\textsuperscript{+}) & 0.735 \\
Memory changes (Info\textsuperscript{+}) & 0.732 \\
Memory changes (Relv\textsuperscript{+}) & 0.729 \\
Memory changes (Any) & 0.720 \\
Extracted claim count & 0.713 \\
Content token count & 0.697 \\
TF--IDF sum & 0.692 \\
Token surprisal sum & 0.691 \\
Length (tokens) & 0.687 \\
Novel word count & 0.633 \\
Entity count & 0.439 \\
\bottomrule
\end{tabular}
\caption{Same as Table~\ref{tab:feat_corr_human} but using the \texttt{CIG} soft label predicted by \textbf{GPT-5} as the target variable (reported for robustness in the appendix).}
\label{tab:feat_corr_gpt5}
\end{table}

\section{Pipeline specification} \label{appendix:pipeline_spec}

Transcript segmentation is performed using GPT-5 with "minimal" reasoning effort settings (accessed in August, 2025). Claim extraction is carried out using GPT-5-mini, also with "minimal" reasoning effort settings (accessed in August, 2025).

For the retrieval layer, extracted claims are embedded using \texttt{Qwen/Qwen3-Embedding-0.6B}, a dense embedding model. The embedding vectors are stored in Qdrant, which serves as the vector database for retrieval.

Memory consolidation is performed using GPT-5-mini with "minimal" reasoning effort (accessed in August, 2025). Segment-level memory summarization is then conducted using GPT-5 with default inference settings (accessed in August, 2025). Finally, CIG ratings are produced using GPT-5 with default inference settings (accessed in August, 2025), with additional models evaluated in ablation experiments.

Local-model experiments are run on the Spartan cluster using \texttt{Qwen/Qwen3-4B-Instruct-2507} on a single NVIDIA A100 GPU.

\section{Plain language statement (anonymised)}\label{appendix:pls}

\noindent\textbf{Project overview.}
This research studies how community audiences perceive \textbf{informativeness} in public deliberative conversations (e.g., debates, community meetings, and other public forums). \par

\noindent\textbf{What you will do.}
You will complete a short training/pilot task, then (optionally) additional annotation tasks. In each task, you will read a debate topic, brief background information, and short segments of conversation, and rate each target utterance on informativeness-related dimensions (e.g., Novelty, Relevance, and Scope of implications) using 1--4 scales. \par

\noindent\textbf{Time commitment.}
Each task is designed to be completed within a bounded time window; a pilot phase is used to estimate average completion time and set payments accordingly. \par

\noindent\textbf{Benefits.}
Your participation contributes to research on how people judge informative communication in deliberation, supporting the development of tools for more constructive and effective public conversations. \par

\noindent\textbf{Payment.}
You will be compensated in line with platform guidelines and to meet or exceed applicable minimum-wage expectations given typical task duration. \par

\noindent\textbf{Risks / discomfort.}
No significant risks are anticipated. However, some segments may contain sensitive, controversial, or emotionally charged content. You may stop at any time if you feel uncomfortable. \par

\noindent\textbf{Voluntary participation / opt-out.}
Participation is voluntary. You may withdraw at any time without negative consequences. Each segment includes sufficient topic/context information beforehand, and you may opt out of rating any specific item if desired. \par

\noindent\textbf{Privacy / data handling.}
No direct personal identifiers are collected by the research team beyond platform-managed participant IDs. Data will be de-identified prior to release; only anonymised annotations and derived research outputs will be shared publicly. \par

\noindent\textbf{Results sharing.}
Aggregate findings and anonymised resources (e.g., code and de-identified annotations) may be released via academic publications and public repositories. \par

\noindent\textbf{Contacts and complaints.}
For anonymous review, institutional identifiers and direct contact details are withheld. A non-identifying institutional ethics-review process approved the study; in the camera-ready version, the final manuscript will include the responsible office/contact channel for participant concerns. \par

\paragraph{Latency.}
To assess the practical cost of our memory-based pipeline, we report latency for (i) segment-level rating under a reused \textsc{memory} context and (ii) per-utterance memory updates. Table~\ref{tab:rating_latency} shows end-to-end inference time for rating a segment when the memory context is carried over across three consecutive segments. Table~\ref{tab:memory_ops_latency} breaks down the memory update cost into claim extraction, retrieval+matching, and integration for 49 utterance updates. Across models, retrieval+matching dominates the memory update runtime, while integration is negligible; in rating, smaller/faster models substantially reduce end-to-end latency.

\begin{table}[H]
\centering
\footnotesize
\setlength{\tabcolsep}{6pt}
\renewcommand{\arraystretch}{1.15}
\begin{tabular}{l r r r}
\toprule
\textbf{Model} & \textbf{Mean} & \textbf{Std} & \textbf{$N$} \\
\midrule
OpenAI GPT-5      & 12.129 & 5.128 & 49 \\
OpenAI GPT-5-mini & 13.964 & 1.106 & 49 \\
Gemini 2.5 Pro    &  8.829 & 1.720 & 49 \\
Gemini 2.5 Flash  &  5.147 & 1.535 & 49 \\
\bottomrule
\end{tabular}
\caption{Rating inference latency in memory mode (seconds), measured on 3 consecutive segments. ChatGPT uses \texttt{reasoning\_effort=minimal} and Gemini uses \texttt{thinking\_budget=128}.}
\label{tab:rating_latency}
\end{table}

\begin{table}[H]
\centering
\footnotesize
\setlength{\tabcolsep}{4pt}
\renewcommand{\arraystretch}{1.15}
\begin{tabular}{p{0.23\columnwidth} p{0.27\columnwidth} r r r}
\toprule
\textbf{Backend} & \textbf{Operation} & \textbf{Mean} & \textbf{Std} & \textbf{$N$} \\
\midrule
GPT-5-mini & Claim extraction     & 2.536 & 1.843 & 49 \\
GPT-5-mini & Retrieval + matching & 5.046 & 5.331 & 49 \\
GPT-5-mini & Memory integration   & 0.067 & 0.124 & 49 \\
GPT-5-mini & Total                & 7.650 & 7.147 & 49 \\
\midrule
Gemini-2.5-Flash & Claim extraction     & 2.791 & 1.752 & 49 \\
Gemini-2.5-Flash & Retrieval + matching & 3.566 & 3.474 & 49 \\
Gemini-2.5-Flash & Memory integration   & 0.057 & 0.144 & 49 \\
Gemini-2.5-Flash & Total                & 6.414 & 5.185 & 49 \\
\bottomrule
\end{tabular}
\caption{Memory operation latency by backend (seconds), measured on 3 consecutive segments (49 utterances total). ChatGPT uses \texttt{reasoning\_effort=minimal} and Gemini uses \texttt{thinking\_budget=128}.}
\label{tab:memory_ops_latency}
\end{table}

\end{document}